\documentclass{article}
\usepackage{graphicx} 
\usepackage{authblk}
\usepackage{subcaption}
\usepackage{caption}
\usepackage[onehalfspacing]{setspace}
\usepackage{float}
\usepackage{url}
\usepackage{makecell}
\usepackage{tabularray}
\usepackage{tabularx}
\usepackage{booktabs}
\usepackage{siunitx}
\usepackage[hidelinks]{hyperref}
\usepackage[margin=2cm]{geometry}
\usepackage{amsmath, amsthm, amssymb}
\usepackage{bm}
\usepackage{minted}
\usepackage{booktabs}
\usepackage{multirow}
\usepackage{xurl}
\newcommand\rurl[1]{%
  \href{https://#1}{\nolinkurl{#1}}%
}
\usepackage{algorithm}
\usepackage{algpseudocode}
\usepackage[numbers,sort&compress]{natbib}
\usepackage{siunitx}
\usepackage{xr}    
\newcommand{\Tcond}{T^{\mathrm{cond}}}

\newcommand{\Pset}{P^\mathrm{set}}
\newcommand{\pdrop}{\Delta P} 
\newcommand{\Qloss}{\dot{Q}_{\mathrm{loss}}}
\newcommand{\Qlossreb}{\dot{Q}_{\mathrm{loss}}^{\mathrm{reboiler}}}

\newcommand{\Qset}{\dot{Q}^{\mathrm{set}}_{\mathrm{in}}}
\newcommand{\Qinreb}{\dot{Q}_{\mathrm{in}}}

\newcommand{\Umat}{U_{\mathrm{mat}}}
\newcommand{\cp}{c_{p}}
\newcommand{\Tref}{T^{\mathrm{ref}}}
\newcommand{\glass}{\mathrm{glass}}
\newcommand{\steel}{\mathrm{steel}}

\newcommand{\napp}{n^{\mathrm{app}}}

\mathchardef\mhyphen="2D 
\usepackage[normalem]{ulem}
\newcommand{\rev}[1]{\textcolor{black}{#1}}
\newcommand{\secondrev}[1]{\textcolor{black}{#1}}

\DeclareSIUnit\bar{bar}

\title{Automated Batch Distillation Process Simulation for a Large Hybrid Dataset for Deep Anomaly Detection}

\author[1]{Jennifer~Werner\thanks{These authors contributed equally to this work.}}
\author[2]{Justus~Arweiler\protect\footnotemark[1]}
\author[2]{Indra~Jungjohann\protect\footnotemark[1]}
\author[1]{Jochen~Schmid\protect\footnotemark[1]}
\author[2]{Fabian~Jirasek}
\author[2]{Hans~Hasse}
\author[1]{Michael~Bortz}

\affil[1]{Fraunhofer Institute for Industrial Mathematics (ITWM), Kaiserslautern, Germany} 
\affil[2]{Laboratory of Engineering Thermodynamics (LTD), RPTU Kaiserslautern, Germany}

\date{}

\begin{document}

\maketitle

\begin{abstract}
\noindent Anomaly detection (AD) in chemical processes based on deep learning offers significant opportunities but requires large, diverse, and well-annotated training datasets that are rarely available from industrial operations. In a recent work, we introduced a large, fully annotated experimental dataset for batch distillation under normal and anomalous operating conditions. In the present study, we augment this dataset with a corresponding simulation dataset, creating a novel hybrid dataset. The simulation data is generated in an automated workflow with a novel Python-based process simulator that employs a tailored index-reduction strategy for the underlying differential-algebraic equations. Leveraging the rich metadata and structured anomaly annotations of the experimental database, experimental records are automatically translated into simulation scenarios. After calibration to a single reference experiment, the dynamics of the other experiments are well predicted. This enabled the fully automated, consistent generation of time-series data for a large number of experimental runs, covering both normal operation and a wide range of actuator- and control-related anomalies. The resulting hybrid dataset is released openly. From a process simulation perspective, this work demonstrates the automated, consistent simulation of large-scale experimental campaigns, using batch distillation as an example. From a data-driven AD perspective, the hybrid dataset provides a unique basis for simulation-to-experiment style transfer, the generation of pseudo-experimental data, and future research on deep AD methods in chemical process monitoring. \secondrev{In summary, the present work develops and validates a methodology for generating a hybrid experimental-simulation dataset, while the development, training, and evaluation of anomaly detection methods are left to future work.}
\end{abstract}


\section{Introduction}
In chemical engineering, early and reliable anomaly detection (AD) is essential for the safe operation of plants and for preventing failures and accidents. Beyond substantial economic losses, undetected anomalies may pose serious risks to human health and the environment. Industrial chemical processes are monitored by a wide range of sensors that generate diverse multivariate time-series data, which are processed by plant control systems. Although these systems incorporate automated AD functionalities, decision-making still largely relies on the expertise of experienced operating personnel~\cite{Chandola2009, Venkatasubramanian2003, Venkatasubramanian2003a, Venkatasubramanian2003b, Chiang2001}.

Machine learning (ML) provides promising new approaches for enhancing AD in chemical processes~\cite{Hartung2023, wagner2023timesead,Russell2000, Monroy2009, Inoue2017, Chadha2019,  Song2019, Tian2020, Schmidl2022, Wu2024, Darban2024}. However, the development of such methods requires extensive datasets for training and validation and was, until recently, hindered by the lack of suitable data, the only openly available chemical process data being simulation data for the fictitious Tennessee Eastman process~\cite{wagner2023timesead, Downs1993, Rieth2017}. Only recently have datasets from real chemical processes been published, specifically generated to support ML-based AD, including data from a laboratory-scale batch distillation plant~\cite{Arweiler2026} and a continuous distillation mini-plant~\cite{Muraleedharan2025}. Both datasets are accompanied by rich metadata, including comprehensive plant descriptions, and comprise fault-free experiments as well as corresponding anomalous runs with thorough annotations. These datasets were created at universities within the DFG Research Unit “Deep Learning on Sparse Chemical Process Data.” In cooperation with this Research Unit, BASF SE also disclosed data from an industrial process; however, this dataset provides substantially less metadata and annotations than the academic datasets. All three datasets were published in the 2025 NeurIPS Datasets and Benchmarks track~\cite{noBoom}.

Despite this recent progress, the systematic collection of experimental data for the development, training, and validation of AD methods remains highly challenging. In particular, suitably annotated and labeled data from real processes will remain scarce, not only because producing such data is effort-intensive. Industrial processes are designed to operate safely and stably, and deviations from normal operation are both rare and undesirable. Consequently, historical plant data -- if accessible at all -- are typically strongly imbalanced, dominated by normal operation with only a few anomalous events that are often incompletely documented. Moreover, deliberately inducing anomalous operating conditions in experimental or industrial settings is generally unacceptable for safety, regulatory, and economic reasons~\cite{Ji2022, Park2020}. 

Process simulation, therefore, constitutes an attractive complementary data source, as it enables the systematic, reproducible, and scalable generation of process data across a wide range of operating conditions, including controlled variations and well-defined anomalous scenarios. At the same time, simulation data alone are insufficient to fully capture the complexity, variability, and noise characteristics of real process data, limiting their direct applicability to data-driven AD. These considerations motivate hybrid data strategies that combine the coverage and controllability of simulations with the realism of experimental data, aiming to leverage the strengths of both worlds.

We consider batch distillation as a representative example of an instationary chemical process to investigate how dynamic process simulation can support the development of AD methods. Our objective is to generate simulation data for all 119 batch distillation experiments reported in~\cite{Arweiler2026, noBoom} in a fully consistent and automated manner. Achieving this requires developing a dedicated simulation workflow, with a central element: a suitable dynamic process simulation engine.

It is well known that the robust simulation of batch distillation processes is challenging, as it requires the solution of nonlinear differential-algebraic equations with a (differentiation) index of at least $2$~\cite{ascher1998computer, hairer1991ii, kunkel2006differential, Campbell2019}. Accordingly, a wide range of dedicated numerical solution strategies have been developed to solve these equations~\cite{Doherty1978,  VanDongen1985, bachmann1990methods, Aspen, AspenTextbook, cervantes1998, cervantes2000, CERVANTES200041, biegler2002, ragunathan2004, Gruetzmann2006, eckert2008mathematical, lopez2016rigorous, Lopez2016, bortz2019estimating, Mohring2022, qian2023nonlinear, WernerSchmid2025}. In the present work, we build on a recently developed Python-based simulation tool~\cite{WernerSchmid2025}, which establishes and exploits a novel index-reduction approach and enables the robust and fully automated generation of batch distillation simulation data under diverse operating conditions. Compared to the simulation model from~\cite{WernerSchmid2025}, only minor extensions had to be made to reflect the specifics of the concrete batch distillation plant considered here.

We leverage the thus extended simulation engine to generate data that correspond directly to the experimental records reported in~\cite{Arweiler2026, noBoom} and to augment them, in particular by providing information that is inaccessible in the experiments. Together, the experimental and simulation datasets form a unique hybrid dataset that is directly suitable for the development and benchmarking of AD methods. To the best of our knowledge, this dataset represents the largest collection of consistent experimental and simulation data for dynamic chemical processes currently available. 

In the present study, we address the forward problem of generating simulation data from existing experimental data. However, the corresponding inverse problem -- the generation of pseudo-experimental data from simulation data -- is at least equally fascinating. Solving this problem would enable the creation of large, diverse datasets, including those with difficult, unsafe, or undesirable operating conditions, supporting the development of ML-based AD methods. The hybrid dataset introduced here provides a valuable foundation for tackling this inverse problem, for example, through approaches based on style transfer~\cite{Gatys2016, El-Laham2022, Xu2024, nagda2025diffstylets}.

\begin{figure}[H]
 \centering
 \includegraphics[width=0.85\textwidth]{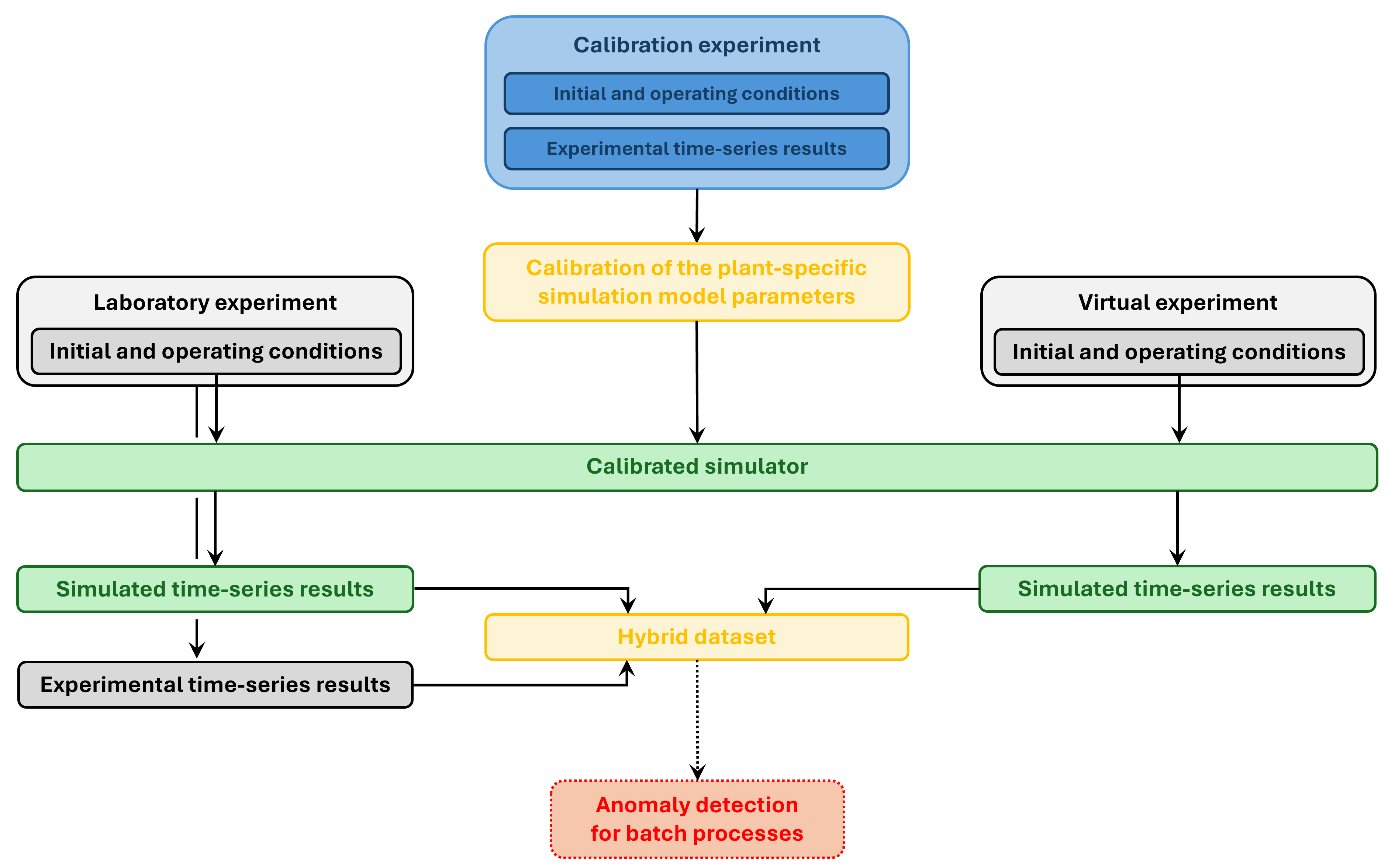}
 \caption{\rev{Schematic of the workflow proposed in this paper. In a first step, the plant-specific parameters of the simulation model are adapted to a single calibration experiment. In a second step, the thus calibrated simulator is used to predict times-series results for other experiments, based only on their initial and operating conditions. In the present paper, experimental time-series data were generated for all simulated experiments (left part of the figure). Additionally, virtual experiments can be simulated for which no experimental data are available (right part of the figure). The simulated time-series results are added to the hybrid experiment-simulation dataset, along with the corresponding experimental time-series data (when available). The resulting hybrid dataset can then be used to train and validate anomaly detection models for batch distillation processes. The present work is focused on the calibration and application to laboratory experiments. The development of anomaly detection methods is future work.}}
 \label{fig:workflow}
\end{figure}

To simulate the experiments from the large dataset~\cite{Arweiler2026, noBoom}, a suitable workflow had to be established \rev{(Figure~\ref{fig:workflow})}. In a first step, the general plant-specific model parameters, such as the number of theoretical stages per meter and thermal capacities, were fixed. For some plant-specific parameters, this was done a priori based on available metadata for the plant. For a subset of plant-specific parameters, an adjustment was made to a single representative \rev{calibration} experiment. In a second step, the other experiments were simulated, \rev{using  the plant-specific parameters that were previously fitted to the calibration experiment}. The experiment-specific parameters were adopted from the database, and no further adjustments were made, so these simulations are predictions. Hence, this corresponds to splitting the experimental dataset into a training set containing a single experiment and a test set containing the remaining 118 experiments. Of those 118 experiments, 4 exhibit liquid-liquid equilibria (LLE), which are not accounted for in the current simulation model. The convergence of these simulations covering different mixtures, feed composition, and operating conditions is not trivial. 
Furthermore, anomalies had to be predicted. Doing this is straightforward for anomalies introduced by perturbing the control parameter setpoints in the simulation. Anomalies stemming from other perturbations, such as the addition of foaming agents or the corruption of sensor data (e.g., due to noise or drift), could not be simulated, as these effects are not covered by the simulation model. In these cases, the simulations covered only the unperturbed phases of the experiment. 

The simulation results obtained in the present work are released openly together with detailed simulation metadata and comprehensive annotations, following the ontology of the experimental database~\cite{Arweiler2026}. Technically, the simulation and experimental datasets are merged into a single hybrid database that contains both data types for 115 experiments, including 36 of the 71 confirmed anomalies described in~\cite{Arweiler2026}. This unique hybrid dataset provides rich opportunities for future research on AD in chemical processes\rev{, tasks that were, however, beyond the scope of our current work and are left for future research.}

\rev{This paper is organized as follows. Section~\ref{sec:apparatus_and_exp_setup} introduces the experimental setup of the laboratory plant used to generate the experimental data. In Section~\ref{sec:simulation_of_the_physical_column}, we present the corresponding simulation model and explain how the model and control parameter values are selected. The model parameters are identified based on a single calibration experiment. In Section~\ref{sec:results}, we compare the simulation results with the experimental data: we first analyze the calibration experiment, then discuss fault-free experiments, and finally evaluate the prediction of anomalous experiments. This section also describes how the simulation data are integrated into the hybrid dataset. Section~\ref{sec:Conclusion} concludes the paper and outlines directions for future work.}

\section{Experimental data} 
\label{sec:apparatus_and_exp_setup}
\begin{figure}[H]
    \centering
    \includegraphics[width=0.75\linewidth]{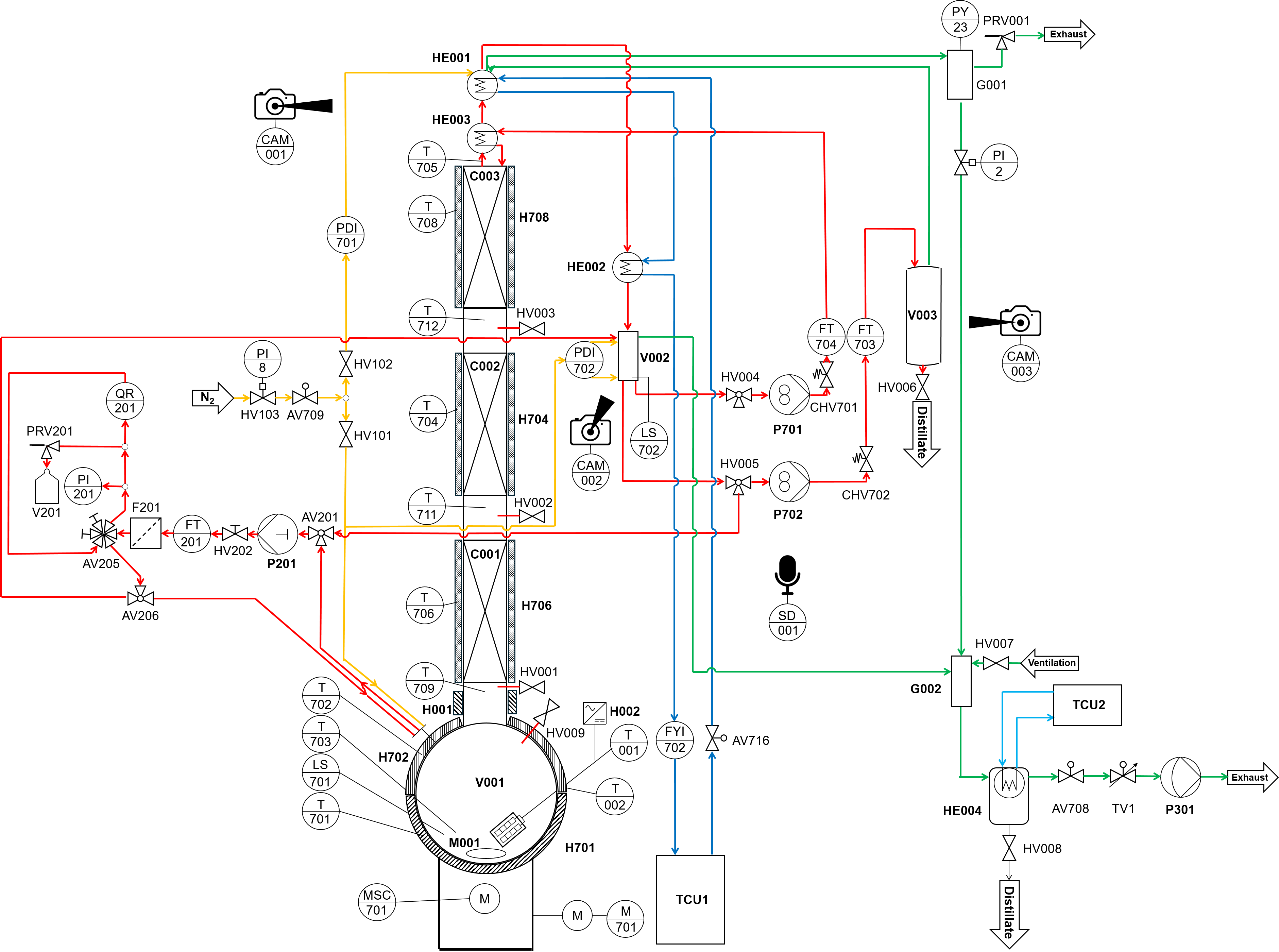}
    \caption{P\&I diagram of the laboratory batch distillation plant. Color code of the lines: product (red), cooling water (dark blue), cooling ethanol (light blue), nitrogen (yellow), and pressure control (green). The equipment list in which the labels are declared is described in~\cite{Arweiler2026}.}
    \label{fig:plant_pid}
\end{figure}

The simulation model developed in this work describes experiments carried out with a batch distillation plant, namely, a modified version of the Iludest LM 2/S glass plant, consisting of a 2 L reboiler vessel and a 1.5 m-high DN30 column section with DX structured packing (Sulzer AG). The piping and instrumentation (P\&I) diagram of the plant is given in Fig.~\ref{fig:plant_pid}. In the experiments, the plant was operated in a static control regime, with fixed constant operating points for pressure, heat supply, and reflux ratio. The initial composition and mass of the feed mixture are known and reported in the database. For a detailed description of the plant and its operation, the reader is referred to~\cite{Arweiler2026}.

In this work, only the operating phase of the experiments is considered; the start-up and shut-down phases are disregarded.
In the batch distillation experiments, anomalies were deliberately introduced by perturbing the process. All anomalies in the distillation processes in the database are described in a comprehensive metadata scheme~\cite{Arweiler2026}, allowing the deduction of the underlying perturbations as adjustments to the time-dependent simulation control parameters. Apart from the start and end time of the applied perturbation, information on the perturbation itself, as well as the affected part of the plant, is available. For the full documentation, the reader is referred to~\cite{Arweiler2026, noBoom}.

\section{Simulation model}
\label{sec:simulation_of_the_physical_column} 

In this section, we introduce the simulation model used to capture the dynamic behavior of the batch distillation plant. Specifically, we spell out the assumptions underlying the model (Section~\ref{subsec:simulation_model_assumptions}) and the model's state variables and control parameters (Section~\ref{subsec:variables}). In Section~\ref{subsec:system_equations}, we then explain how the simulation model from~\cite{WernerSchmid2025} was extended in order to represent the experimental data obtained with the plant described in Section~\ref{sec:apparatus_and_exp_setup}. Finally, we explain how the plant-specific model parameters and the experiment-specific control parameters were set for the various simulations performed to create the hybrid dataset (Section~\ref{subsec:model_parameters}). 

\subsection{Assumptions}
\label{subsec:simulation_model_assumptions}

\begin{figure}[H]
 \centering
 \includegraphics[width=0.65\textwidth]{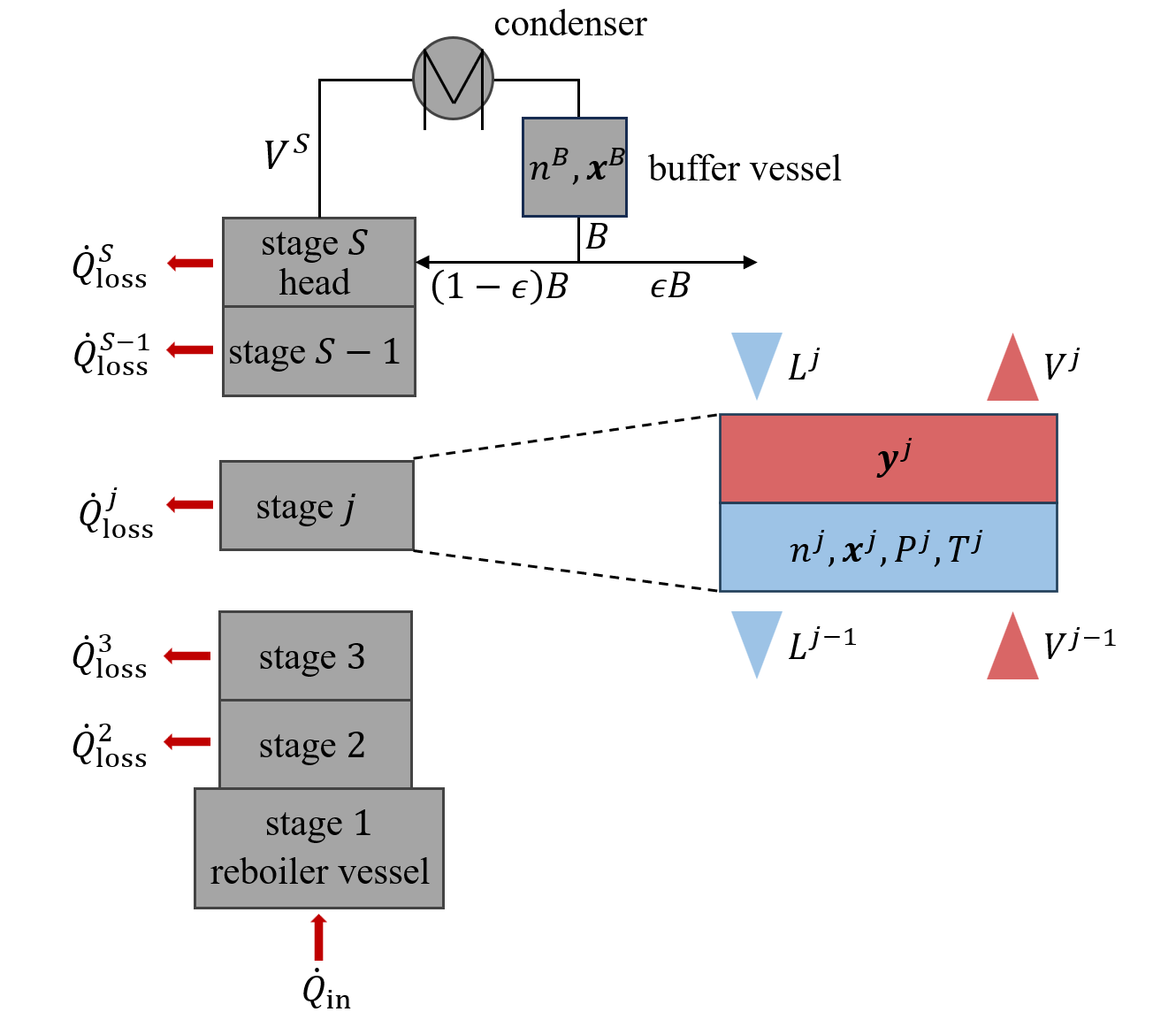}
 \caption{Schematic of the considered batch distillation column model. Symbols are explained in the main text.}
 \label{fig:schematic_simulation_model}
\end{figure}

As most papers on the simulation of batch distillation processes, such as~\cite{Doherty1978,VanDongen1985,cervantes1998, cervantes2000, CERVANTES200041, biegler2002, ragunathan2004, eckert2008mathematical, lopez2016rigorous, bortz2019estimating, Mohring2022, qian2023nonlinear}, we use an equilibrium-stage model for the distillation column and describe the temporal evolution of the  distillation process by means of the mass  and energy balances around the equilibrium stages and by means of phase equilibrium conditions and the summation equations for the individual equilibrium stages. 
Specifically, we consider batch distillation processes in columns with $S\geq2$ stages for multi-component mixtures consisting of $C\geq2$ components. Stage $1$ is the heated reboiler vessel, and stage $S$ is the head, which is connected to a condenser. The condenser, in turn, is connected to a buffer vessel $B$, where the condensed product is gathered, before it is either returned to the column as reflux or withdrawn as distillate. In~\cite{WernerSchmid2025}, we did not consider such a buffer vessel. However, it turned out that incorporating the buffer vessel into the model substantially improved the simulation-experiment fits across the diverse operating and initial conditions covered by the experimental database underlying the present paper. 
A schematic of the batch distillation column model is shown in Figure~\ref{fig:schematic_simulation_model}. 

As in~\cite{WernerSchmid2025}, it is assumed that

\begin{itemize}
    \item[(i)] the vapor holdup on all stages can be neglected, 
    \item[(ii)] there is total condensation in the condenser, 
    \item[(iii)] phases do not disappear, i.e., that exactly two phases are in equilibrium with each other, namely one vapor and one liquid phase, on all stages at all times.
\end{itemize}

Additionally, the mass in the buffer vessel after the condenser is constant in time. Apart from this buffer vessel, we made the following additional changes to the simulation model described in~\cite{WernerSchmid2025}:

\begin{itemize}
    \item[(i)] The internal energy $\Umat^j$ accounting for the energy content of the packing material of stage $j$ and the walls around stage $j$ is given by
    \begin{align} \label{eq:internal-energy-on-stage-j}
        \Umat^j = c^j (m_\steel^j \cp^\steel + m_\glass^j \cp^\glass) (T^j - \Tref)\, , 
    \end{align}
    where $\cp^\steel$ and $\cp^\glass$ are the heat capacities of the employed steel packing and glass walls, respectively, while $m_\steel^j$ and $m_\glass^j$ are the masses of steel packing and the glass walls on or around stage $j$, respectively. Furthermore, $T^j$ is the temperature at stage $j$, $T^\mathrm{ref}$ is a reference temperature for the heat capacities, and the parameters $c^j$ are adjustable correction factors that need to be calibrated to data from the specific plant under consideration. 
    
    \item[(ii)] There are heat losses $\Qloss^j$ from each column stage $j \in \{2, \dots, S\}$ and these are constant in time. 
    
    \item[(iii)] The condenser temperature $\Tcond$ is regulated such that, at all times, it is a constant $\Delta T$ below the head temperature $T^S$. In short,
    \begin{align}
        \Tcond = T^S - \Delta T.
    \end{align} 
    
    \item[(iv)] The pressure on each stage $j$ is described as follows in the simulation model:
    \begin{align}\label{eq:pressure_drop_model}
        P^{j} = P^{S} + (S-j)\frac{\pdrop}{S-1} \qquad (j \in \{1,\dots,S\}),
    \end{align}
    where $P^j$ denotes the pressure on stage $j$ and $\pdrop$ denotes the total pressure drop from the reboiler vessel to the column head.
\end{itemize}

\subsection{Variables} 
\label{subsec:variables}
The state of the considered distillation process is completely described by the following variables:
\begin{itemize}
	\item $n^j$: liquid mole holdup on stage $j \in \{1,\dots,S\}$, 
	\item $n^B$: liquid moles in the buffer vessel,  
	\item $\bm{x}^j = (x^j_1, \dots, x^j_C)$: liquid molar composition on stage $j \in \{1,\dots,S\}$ 
	\item $\bm{x}^B = (x_1^B, \dots, x_C^B)$: liquid molar composition in the buffer vessel, 
	\item $\bm{y}^j = (y_1^j, \dots, y_C^j)$: vapor molar composition on stage $j \in \{1,\dots,S\}$, 
	\item $T^j$: temperature on stage $j \in \{1,\dots,S\}$, 
	\item $L^j$: liquid molar downstream from stage $j+1$ to stage $j$, 
	\item $V^j$: vapor molar upstream from stage $j$ to stage $j+1$, 
	\item $B$: liquid molar stream going out of the buffer vessel,
    \item $D$: molar distillate stream. 
\end{itemize}
Note that $D=\epsilon B$, where $\epsilon$ is the efflux ratio defined below.
Furthermore, the apparatus holdup is defined as follows:
\begin{align}
	\napp := n^1 + \dotsb + n^S + n^B.
\end{align}
The values of the following control (input) parameters as functions of time have to be specified:
\begin{itemize}
	\item $\epsilon$: the efflux ratio 
    \begin{align}
	\epsilon := D/B \in [0,1]
\end{align}
being defined as the share of the liquid stream $B$ that is withdrawn from the column,
	\item $P^S$: the pressure on the head stage $S$, 
	\item $\pdrop$: the pressure drop from the reboiler vessel to the head stage $S$, 
	\item $\Qinreb$: the reboiler heat duty. 
\end{itemize}

\subsection{System equations} \label{subsec:system_equations}

Compared to~\cite{WernerSchmid2025}, the system equations had to be extended as follows:
\begin{itemize}
    \item[(i)] the total and component-wise mass balances around the buffer vessel were added to the system,
    \item[(ii)] the energy balances were extended by the internal energy $\Umat^j$ accounting for the energy content of the packing material and the walls of the corresponding stage, and the heat losses $\Qloss^j$,
    \item[(iii)] an equation ensuring the mass constancy in the buffer vessel was added.
\end{itemize}

This system of equations is spelled out in the Supporting Information.

\subsection{Choice of the model and control parameter values}\label{subsec:model_parameters}
In this section, we discuss how the simulation parameters were set to represent the laboratory batch distillation plant. We distinguish between plant- and experiment-specific parameters. The plant-specific parameters apply only to the plant and are thus the same across all experiments. On the other hand, several parameters are experiment-specific and must be set individually for each experiment according to the available experimental data. This distinction is described in more detail below.

\subsubsection{Plant-specific parameters} \label{subsubsec:global_parameters}
The plant-specific parameters are
\begin{itemize}
    \item the number of stages $S$,
    \item the heat loss $\Qloss^j$ on each stage $j$,
    \item the temperature difference $\Delta T$ between the head and the condenser stage,
    \item the heat capacities of the plant equipment, here the heat capacities $\cp^\steel$ and $\cp^\glass$ of steel and glass, and the masses of the plant equipment, here the masses $m^j_\steel$ and $m^j_\glass$ of steel and glass on each stage $j$,
    \item and the correction parameters $c^j$ in the model~\eqref{eq:internal-energy-on-stage-j}.

\end{itemize}
These parameters were set once and then fixed for all experiments: 

The number of stages is $S = 12$ for each simulation run. This value was determined from measurements of the number of theoretical stages per meter (NTSM) using a binary test-mixture of (ethanol + 2-propanol); details are provided in Supporting Information.

The column of the laboratory distillation plant is insulated (heating elements are installed along the column to minimize heat losses), while the column head is not insulated. Accordingly, in the simulation model, we assumed no heat losses in the column stages and a small heat loss of \SI{2}{\watt} at the top of the column:
\begin{align}
    \Qloss^2=\cdots=\Qloss^{S-1} := \SI{0}{\watt} \qquad \mathrm{and} \qquad \Qloss^S := \SI{2}{\watt}.
\end{align}

Concerning the temperature offset $\Delta T$ between the head and the condenser stage (Assumption 3 (iii)), we assumed
\begin{align}
    \Delta T := \SI{20}{\kelvin}. 
\end{align}

To model the thermal inertia of the plant as in~\eqref{eq:internal-energy-on-stage-j}, masses and heat capacities of the plant equipment must be known. The reboiler vessel, including heating elements and insulation, has a mass of $m^1_{\glass}=$~\SI{1.25}{\kilogram}, the insulation and the submerged heating rod account for a combined mass of $m^1_{\steel}=$~\SI{0.25}{\kilogram}, which was determined gravimetrically. In contrast, the masses of glass and structured packing on the theoretical separation stages cannot be measured gravimetrically and were estimated to be $m^j_{\glass}=$~\SI{0.2}{\kilogram} and $m^j_{\steel}$~\SI{0.058}{\kilogram} for all $j \in \{2,\dots,S\}$. 
The glass material used in the plant is borosilicate-glass with a heat capacity of $\cp^\glass=$~\SI{830.0}{\joule \kilogram^{-1} \kelvin^{-1}}~\cite{HeatCapacity_Borosilicate-glass}. The structured packing, as well as the heating rod, is made from alloy 1.4404 316L with a heat capacity of $\cp^\steel=$~\SI{510.79}{\joule \kilogram^{-1} \kelvin^{-1}}~\cite{HeatCapacity_14404stainlesssteel}. Furthermore, we set the reference temperature $\Tref=$~\SI{273.15}{\kelvin}.

To obtain the correction parameters $c^j$, cf.~Eq.~(\ref{eq:internal-energy-on-stage-j}), an adjustment to a single representative experiment was performed. Calibrating them to a single experiment (see Table~\ref{tab:experiment_Ids}), we found the values
\begin{align}
    c^1 := 1, c^2=2.5, \dots, c^S := 2.5
\end{align}
to be appropriate for the plant considered here.
\rev{The correction parameters $c^j$ were chosen manually. Starting from the default value of the correction parameters $c^j=1, j=1, ..., S$, the parameters were adjusted stepwise. For each candidate parameter set we ran simulations and compared the outputs with the corresponding experimental data. Based on overall shape, timing, and main trends of the experimental data and the sum of squared errors between experiments and simulations, a suitable parameter set could be identified. This iterative process was continued until further parameter changes did not produce a meaningful reduction of the discrepancies. The heat loss at the column top was set in a similar manner. The final parameter set was subsequently tested against independent experimental datasets to ensure that the calibrated model retained predictive capability beyond the calibration conditions.}

\subsubsection{Experiment-specific parameters} \label{subsubsec:experiment_specific_parameters}
Each experiment was conducted at a unique operating point; therefore, the experiment-specific parameters must be set individually. These parameters were determined partly from the plant control settings and partly from the measurement data.
In our case, the experiment-specific parameters are
\begin{itemize}
    \item the pressure at the column head $P^S$ and the pressure drop $\Delta P$,
    \item the heat duty $\Qinreb$ applied to the reboiler,
    \item and the efflux ratio $\epsilon$.
\end{itemize}

The pressure $P^S$ for each experiment was set equal to the setpoint head pressure $\Pset$, and the pressure drop $\Delta P$ was determined from the experimentally observed pressure drop. The pressures $P^j, j \in \{1, \dots, S-1\}$ were then determined by the simple linear pressure drop model given in Eq.~\eqref{eq:pressure_drop_model}.

The reboiler heat duty $\Qinreb$ for each experiment was determined according to
\begin{equation}
    \Qinreb = \Qset - \Qlossreb 
\end{equation}
where $\Qset$ is the setpoint heat flux and $\Qlossreb$ is the reboiler heat loss, which is modeled as explained in the Supporting Information.

Consistent with the constant reflux-ratio control strategy implemented in the laboratory batch distillation plant (Section~\ref{sec:apparatus_and_exp_setup}), the efflux ratio $\epsilon$ was set to a constant value in all simulations from Section~\ref{sec:results} (except when it was perturbed as in Section~\ref{subsubsec:reflux_anomaly_prediction}, of course). This value was determined by averaging the experimentally observed reflux-ratio values over the experiment, \rev{excluding all anomalous time points.}

\section{Results and discussion} 
\label{sec:results}

\subsection{Overview}
\label{subsec:results_overview}
Using the general simulation model introduced in Section~\ref{sec:simulation_of_the_physical_column}, we simulated all experiments performed with the batch distillation plant described in Section~\ref{sec:apparatus_and_exp_setup}, except for runs exhibiting liquid–liquid equilibria (LLE). This section presents five representative cases: the calibration experiment (Section~\ref{subsec:results_calibration}), one fault-free experiment (Section \ref{subsec:normal_prediction}), and three anomalous experiments (Section~\ref{subsec:results_prediction}). Section~\ref{subsec:hybrid_dataset} describes the hybrid dataset established in this work.

The experimental batch distillation database~\cite{Arweiler2026, noBoom} documents the anomalies and their physical causes in detail. As outlined in Section~\ref{sec:apparatus_and_exp_setup}, anomalies can be represented in the model as time-dependent modifications of control signals, provided their causes are known and correspond to actuator setpoint changes. Perturbations in pressure, heating or cooling duties, condenser operation, efflux ratio, leakage, or material withdrawal can thus be mimicked.

In contrast, anomalies affecting fluid dynamics, mixture properties (e.g., addition of foaming agents), or the control regime (e.g., sensor noise or drift) cannot be mapped onto model control parameters. These cases are therefore not included. 

For the anomalies considered here, the model requires only (i) the start time of the perturbation, (ii) the time of its removal, and (iii) the modified setpoint value. Transitions were implemented as linear ramps to ensure numerical stability and to reflect experimentally observed rates of change. The three anomalous runs discussed here involve perturbations in the efflux ratio, heat duty, and system pressure.

For the experiments discussed here, the identifiers, experiment-specific parameters, and initial conditions at simulation start time $t_\mathrm{start}=0$ are listed in Table~\ref{tab:experiment_Ids}. Results for four additional fault-free experiments, \rev{including one experiment conducted with the ternary system (acetone + 1-butanol + methanol),} are given in the Supporting Information. The corresponding information for the entire body of the experiments is reported in the published hybrid dataset~\cite{MainDataset}, including also detailed comparisons between the simulations and the experimental results.

\begin{table}[h]
    \centering
    \caption{Overview of the experiments discussed here: Experiment-specific parameters, initial conditions \rev{(composition of the reboiler vessel and apparatus holdup)} $\bm{x}^1(t_\mathrm{start}=0)$ and $n^\mathrm{app}(t_\mathrm{start}=0)$, and assignment to the database. All experiments are from the folder containing the results for the system \rev{(1-butanol + 2-propanol} + water). Entries marked with a dagger ($^\dag$) were perturbed in the simulation.}
    \begin{tabular}{lcccccccc}
    \hline
    Case & ID & $\Qinreb$ / W & $P^S$ / Pa & $\pdrop$ / Pa & $\epsilon$  & $\bm{x}^1(0)$ / mol/mol & $n^\mathrm{app}(0)$ / mol \\
    \hline \hline
    Calibration        & 1 & 230.71   & 70000     & 93.0 & 0.30       & $(0.429,0.43,0.141)$  & 24.62 \\ 
    Fault-free       & 2 & 79.77    & 70000     & 75.0 & 0.44       & $(0.454,0.394,0.152)$ & 17.85 \\
    Reflux-ratio perturbation    & 3  & 79.86    & 60000     & 64.0 & 0.30$^\dag$& $(0.348,0.433,0.219)$ & 18.38 \\
    Heat-duty perturbation & 4  & 77.51$^\dag$ & 70000    & 72.0 & 0.45    & $(0.442,0.431,0.127)$ & 24.61 \\
    Pressure perturbation  & 5  & 183.90    & 50000$^\dag$& 82.0 & 0.17    & $(0.276,0.482,0.242)$ & 19.59 \\
    \hline
    \end{tabular}
    \label{tab:experiment_Ids}
\end{table}

\subsection{Calibration experiment}
\label{subsec:results_calibration}
\begin{figure}[H]
    \centering
        \begin{subfigure}[t]{0.95\textwidth}
        \centering
        \includegraphics[width=\linewidth]{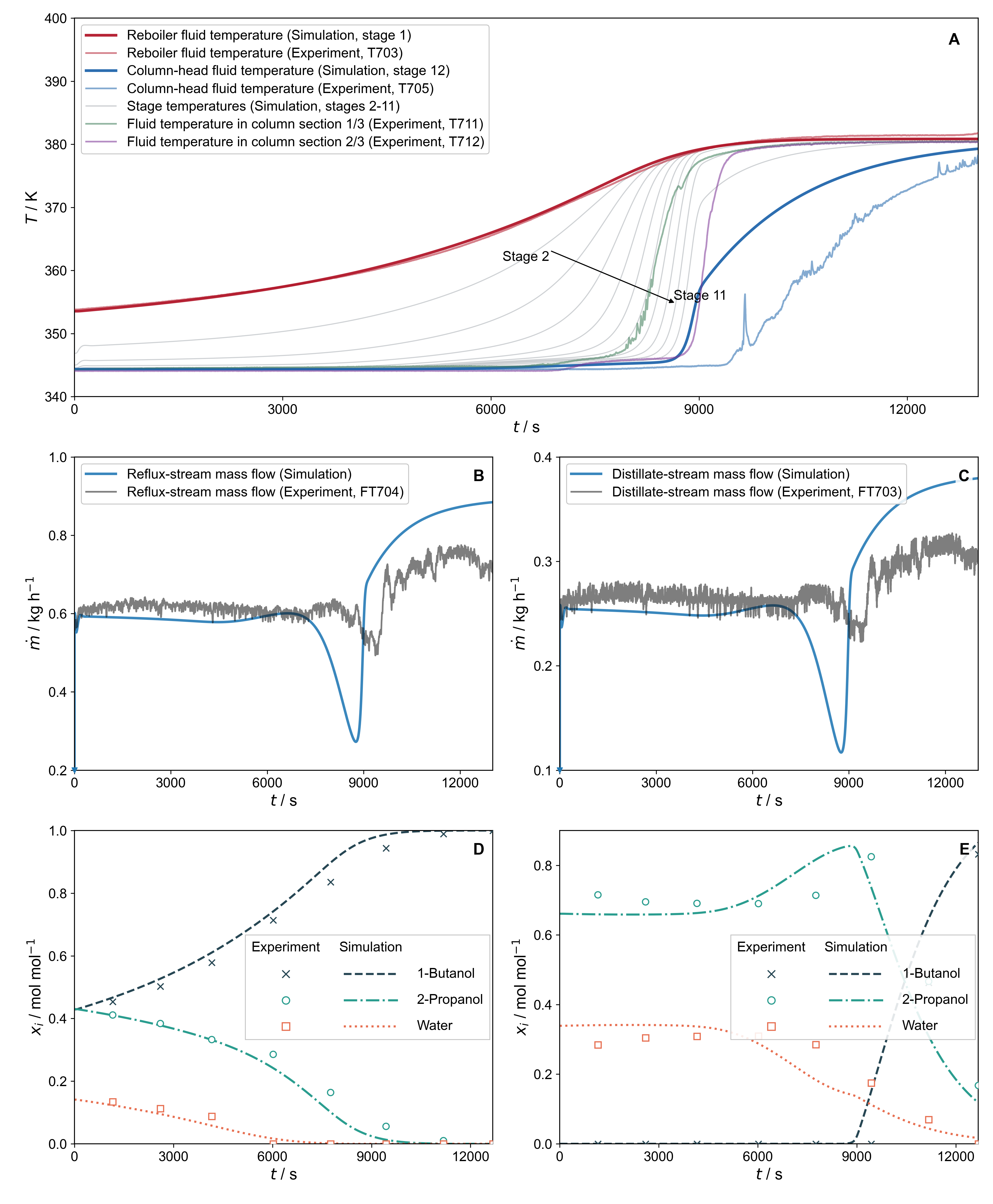}
    \end{subfigure}
    \caption{\rev{Comparison of measurements and simulation results for the fault-free calibration experiment (ID~1). Panel A shows the temperatures of all column stages. In the experiment, one temperature was recorded per column section, with the temperature sensors located above the respective sections. Panels B and C show the reflux and distillate mass flows. Panels D and E compare the liquid-phase composition over time in the reboiler vessel and condensate buffer vessel, respectively.}}
    \label{fig:simexp_normal_2024_09_13_V2_combined}
\end{figure}

The batch distillation model was calibrated using normal-operation data from a single experiment with the ternary system (1-butanol + 2-propanol + water). The chosen operating point was a moderate one with respect to pressure, reflux ratio, heat duty, and feed composition, avoiding extreme conditions. \rev{Although the calibration experiment has a higher heat duty than the other experiments shown here, its heat duty remains well within the operating range of the plant and is not close to the technical upper limit of \SI{750}{\watt}.} The experiment was run until the reboiler vessel was depleted and the plant was shut down. 
\rev{Figure~\ref{fig:simexp_normal_2024_09_13_V2_combined} compares the simulation results with the corresponding calibration experiment (ID~1). The simulation model reproduces the main process behavior well throughout the entire duration of the experiment.
The simulated temperatures in the reboiler vessel and along the column agree reasonably well with the experimental trends and absolute values, as shown in panel~A of Figure~\ref{fig:simexp_normal_2024_09_13_V2_combined}. However, the dynamic batch-distillation behavior progresses faster in the simulation, with changes propagating earlier along the column toward the top.
The simulated reflux and distillate mass flows, shown in panels~B and C of Figure~\ref{fig:simexp_normal_2024_09_13_V2_combined}, reproduce the overall experimental trends. The transition to the next-higher-boiling component is accompanied by a temporary decrease in the mass flows, which is more pronounced in the simulation than in the experiment. 
This deviation is consistent with the earlier progression of the batch dynamics and is mainly attributed to the simplified representation of fluid dynamics and thermal inertia in the column. Further deviations in timing and absolute values can be attributed to the simplified treatment of heat losses and holdup, the selected heat duty, and uncertainty in the flow measurements.
The evolution of the liquid-phase composition in the reboiler vessel and distillate buffer vessel is described correctly overall (panels~D and E of Figure~\ref{fig:simexp_normal_2024_09_13_V2_combined}).}

\subsection{Examples for the prediction of fault-free experiments}
\label{subsec:normal_prediction}
\begin{figure}[H]
        \begin{subfigure}[t]{0.95\textwidth}
        \centering
        \includegraphics[width=\linewidth]{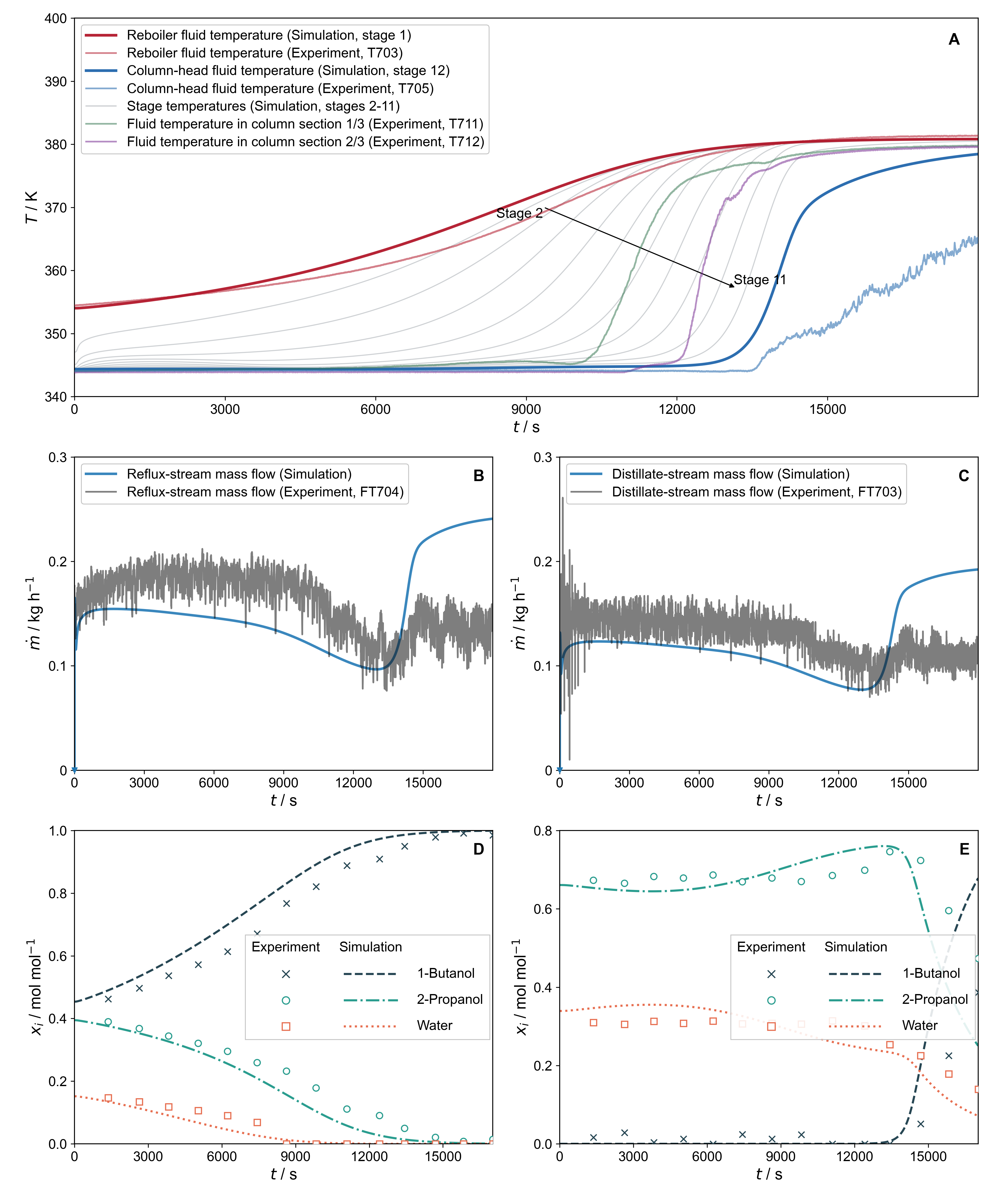}
    \end{subfigure}
    \caption{\rev{Comparison of measurements and simulation results for the fault-free validation experiment (ID~2). The plant-specific model parameters were determined from the calibration experiment (ID~1) and were not adjusted further. Panel A shows the temperatures of all column stages. Panels B and C show the reflux and distillate mass flows. Panels D and E compare the liquid-phase composition over time in the reboiler vessel and condensate buffer vessel, respectively.}}
    \label{fig:simexp_normal_2024_08_19_combined}
\end{figure}

Figure~\ref{fig:simexp_normal_2024_08_19_combined} compares the model predictions with experimental data for \rev{the fault-free validation experiment (ID~2) to assess the application of the calibrated simulation model to different operating conditions.} The conditions differ markedly from those of the calibration experiment; in particular, the heat duty was significantly lower, cf. Table~\ref{tab:experiment_Ids}. No adjustments to the plant-specific parameters determined for the calibration experiment (cf.~Section~\ref{subsec:results_calibration}) were made; the simulation results shown in Figure~\ref{fig:simexp_normal_2024_08_19_combined} are true predictions. The dynamic behavior is well captured, particularly with respect to the composition \rev{as shown in panels~D and E of Figure~\ref{fig:simexp_normal_2024_08_19_combined}}. Nevertheless, some quantitative discrepancies remain, especially towards the end of the experiment. Overall, the quality of the predictions is comparable to that of the calibration experiment, indicating the robustness of the model. \rev{For the anomalous experiments discussed in the following section (ID~3--5), no further adjustments were made to the plant-specific parameters determined from the calibration experiment (cf.~Section~\ref{subsec:results_calibration}). Together with the results for the fault-free experiments reported in the Supporting Information (ID~6--9), this supports the applicability of the simulation model over a broad range of initial compositions and mixture properties.}

\subsection{Examples for the prediction of anomalous experiments}
\label{subsec:results_prediction}
\subsubsection{Anomaly induced by \rev{reflux}-ratio perturbation} \label{subsubsec:reflux_anomaly_prediction}

\begin{figure}[H]
    \centering
        \begin{subfigure}[t]{0.95\textwidth}
        \centering
        \includegraphics[width=\linewidth]{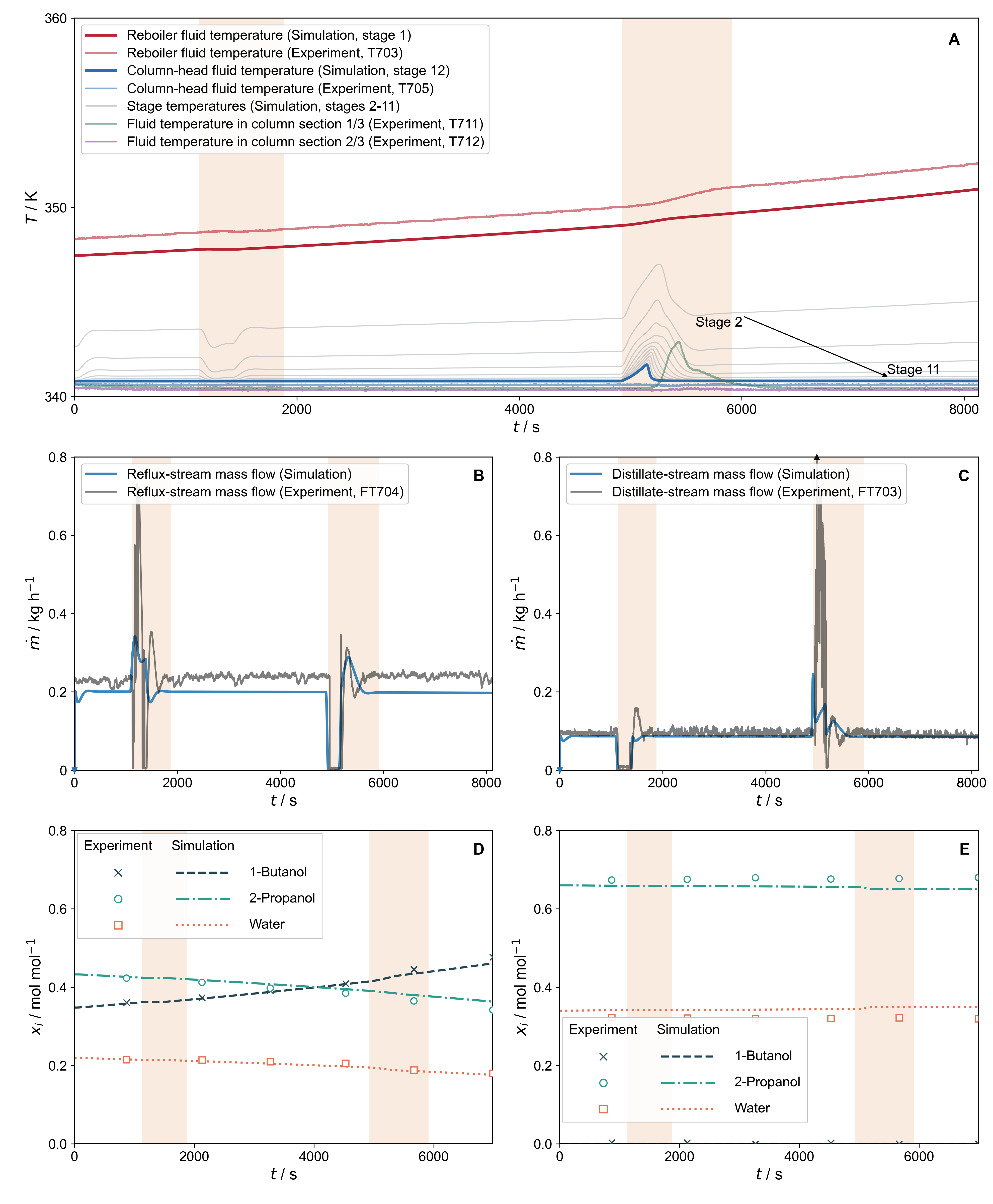}
    \end{subfigure}
    \caption{\rev{Comparison of measurements and simulation results for the anomalous experiment ID~3 showing an anomaly induced by a reflux-ratio perturbation. Shaded areas mark the anomalous time window from the introduction of the perturbation to the end of the visible effect in the measurements. Panel A shows the temperatures of all column stages. Panels B and C show the reflux and distillate mass flows and panels D and E the liquid-phase composition over time in the reboiler vessel and condensate buffer vessel, respectively.}}
    \label{fig:simexp_normal_2024_07_18_combined}
\end{figure}

\rev{We now discuss the anomalous experiment ID~3 in which the reflux ratio was deliberately perturbed by temporary step changes. In the first perturbation, the reflux ratio was increased, whereas in the second, the reflux ration was decreased. Figure~\ref{fig:simexp_normal_2024_07_18_combined} compares the simulated results with the corresponding measurements, including temperatures in the reboiler vessel and along the column, liquid-phase composition in the reboiler and buffer vessels, and the distillate and reflux mass flows.
The model reproduces the unperturbed operation well overall and predicts the process response to both perturbations with reasonable accuracy. In particular, the recovery of the mass flows after resetting the reflux ratio to its normal value is well described (see panels~B and C of Figure~\ref{fig:simexp_normal_2024_07_18_combined}), although the amplitude of the response is underestimated. This deviation is likely related to the control behavior of the reflux and distillate pumps.
The effect of the perturbation on the reboiler temperature is also predicted correctly as it can be seen in panel~A of Figure~\ref{fig:simexp_normal_2024_07_18_combined}, although the response is subtle. In the stage temperatures, the perturbation is only visible at the sensor above the first column section and only for the second step change. The first perturbation is not visible in the experimental stage-temperature data. This limited observability is attributed to operation near the lower capacity limit of the column, where low vapor and liquid loads can lead to nonuniform liquid downflow, such as rivulet formation, which is not fully captured by the temperature sensors.
A similar limitation applies to the composition measurements seen in panels~D and E of Figure~\ref{fig:simexp_normal_2024_07_18_combined}. While the temporal resolution of the experimental composition data is too low to resolve the local anomaly predicted by the simulation, the overall composition profiles are reproduced well.} 

\subsubsection{Anomaly induced by heat\rev{-}duty perturbation} \label{subsubsec:heatduty_anomaly_prediction}

\begin{figure}[H]
    \begin{subfigure}[t]{0.95\textwidth}
    \centering
    \includegraphics[width=\linewidth]{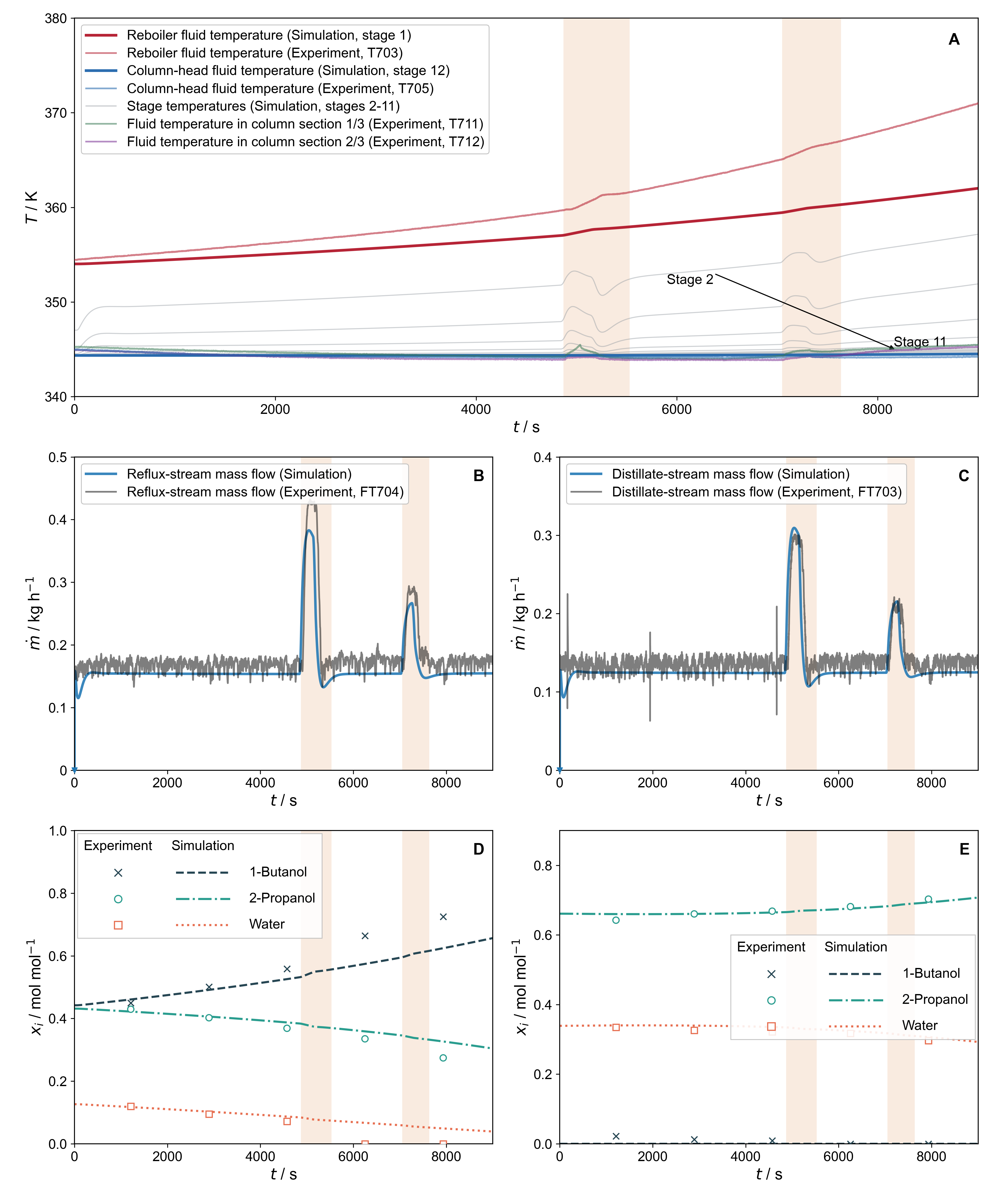}
    \end{subfigure}
    \caption{\rev{Comparison of experiment and simulation results for the anomalous experiment ID~4 with heat-duty perturbations. Shaded areas mark the anomalous time window from the introduction of the perturbation to the end of the visible effect in the measurements. Panel A shows the temperatures of all column stages. Panels B and C show the reflux and distillate mass flows, respectively. Panels D and E compares the liquid-phase composition over time in the reboiler vessel and condensate buffer vessel, respectively.}}
    \label{fig:simexp_normal_2024_09_12_combined}
\end{figure}

\rev{As a second anomalous example, we consider experiment ID~4 in which the heat input to the reboiler was temporarily changed. The first perturbation consists of a step-like increase in heat duty to a higher constant value, whereas the second perturbation decreases the heat input to a lower constant value.
Figure~\ref{fig:simexp_normal_2024_09_12_combined} compares the simulated and experimental responses of the perturbation.
The simulation describes the experiment with good quantitative accuracy. The main deviation is observed in the reboiler-temperature evolution (see panel~A of Figure~\ref{fig:simexp_normal_2024_09_12_combined}), where the experimental profiles increase more steeply than the simulated ones, which may result from a small underestimation of the derived heat duty, together with the simplified description of heat losses and holdup.
The simulated temperatures along the stages further matches the experimental response. The sensor above the first column section shows the clearest experimental effect and matches the predicted behavior well. Unlike in the reflux-ratio perturbation experiment, the other stage temperatures also show a response to both perturbations, although the effect is weaker. This is expected, as increasing the heat duty directly affects the amount of uprising vapor.
The simulation reproduces the reflux and distillate mass flows well, as shown in panels~B and C of Figure~\ref{fig:simexp_normal_2024_09_12_combined}. Both the shape and amplitude of the experimentally observed response are predicted accurately.
This shows that the model captures the dominant effect of the altered heat input on the fluid dynamics of the column.
The composition measurements show the same limitations discussed above (see panels~D and E of Figure~\ref{fig:simexp_normal_2024_09_12_combined}).
Overall, however, the heat-duty perturbation experiment is described well, with strong agreement observed for the mass-flow response.}

\subsubsection{Anomaly induced by \rev{system-}pressure perturbation} \label{subsubsec:pressure_anomaly_prediction}

\begin{figure}[H]
    \begin{subfigure}[t]{0.95\textwidth}
    \centering
    \includegraphics[width=\linewidth]{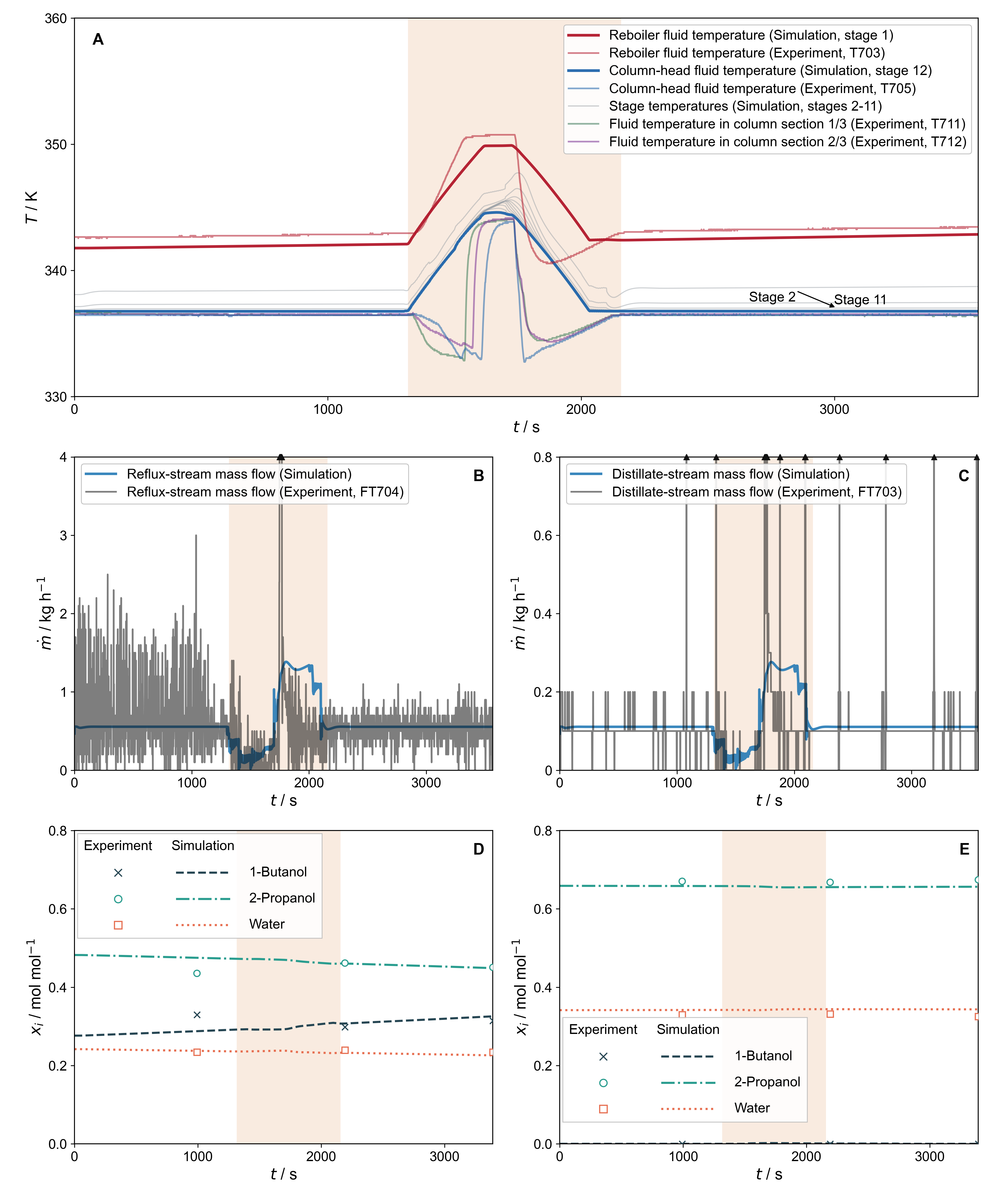}
    \end{subfigure}
    \caption{\rev{Comparison of measurements and simulation results for the anomalous experiment ID~5 with a system-pressure perturbation. Shaded areas indicate the anomalous time window from the introduction of the perturbation to the end of its visible effect in the experiment. Panel A shows the temperatures of all column stages. Panels B and C show the reflux and distillate mass flows, respectively. Spikes in the experimental distillate mass flow are attributed to gas bubbles passing through the sensor. Panels D and E compares the liquid-phase composition over time in the reboiler vessel and condensate buffer vessel, respectively.}}
    \label{fig:simexp_normal_2023_10_13_combined}
\end{figure}

\rev{The third perturbation scenario considers the experiment ID~5 in which the system pressure was temporarily increased. Figure~\ref{fig:simexp_normal_2023_10_13_combined} compares the simulated results with the corresponding experiment, including the stage temperatures, liquid-phase composition, and distillate and reflux mass flows.
While the model captures the unperturbed operation well overall, larger deviations arise during the perturbation and subsequent recovery phase compared with the previously discussed perturbation examples. This is mainly because the pressure controller itself was not included in the model. Instead, the pressure change was imposed in a simplified form (cf. panel A of Figure~\ref{fig:simexp_normal_2023_10_13_combined}). Still, the simulation qualitatively agrees with the experiments, but differs in the detailed shapes of the curves, as shown in Figure~\ref{fig:simexp_normal_2023_10_13_combined}. The mass-flow response also is reproduced qualitatively, and the direction of the perturbation effect is reflected in the simulated process behavior as shown in panels~B and C of Figure~\ref{fig:simexp_normal_2023_10_13_combined}. Spikes seen in the experimental distillate mass flow are attributed to gas bubbles passing through the sensor. The largest deviations occur in the temperature predictions as shown in panel~A of Figure~\ref{fig:simexp_normal_2023_10_13_combined}, which are particularly sensitive to the pressure dynamics and controller response. The composition measurements show the same limitations as in the previous perturbation cases (see panels~D and E of Figure~\ref{fig:simexp_normal_2023_10_13_combined}). 
Overall, the model describes the main process response to the system-pressure perturbation well.}

\subsection{Simulation data within the hybrid dataset}
\label{subsec:hybrid_dataset}
\begin{figure}[H]
   \centering
   \includegraphics[width=0.95\linewidth]{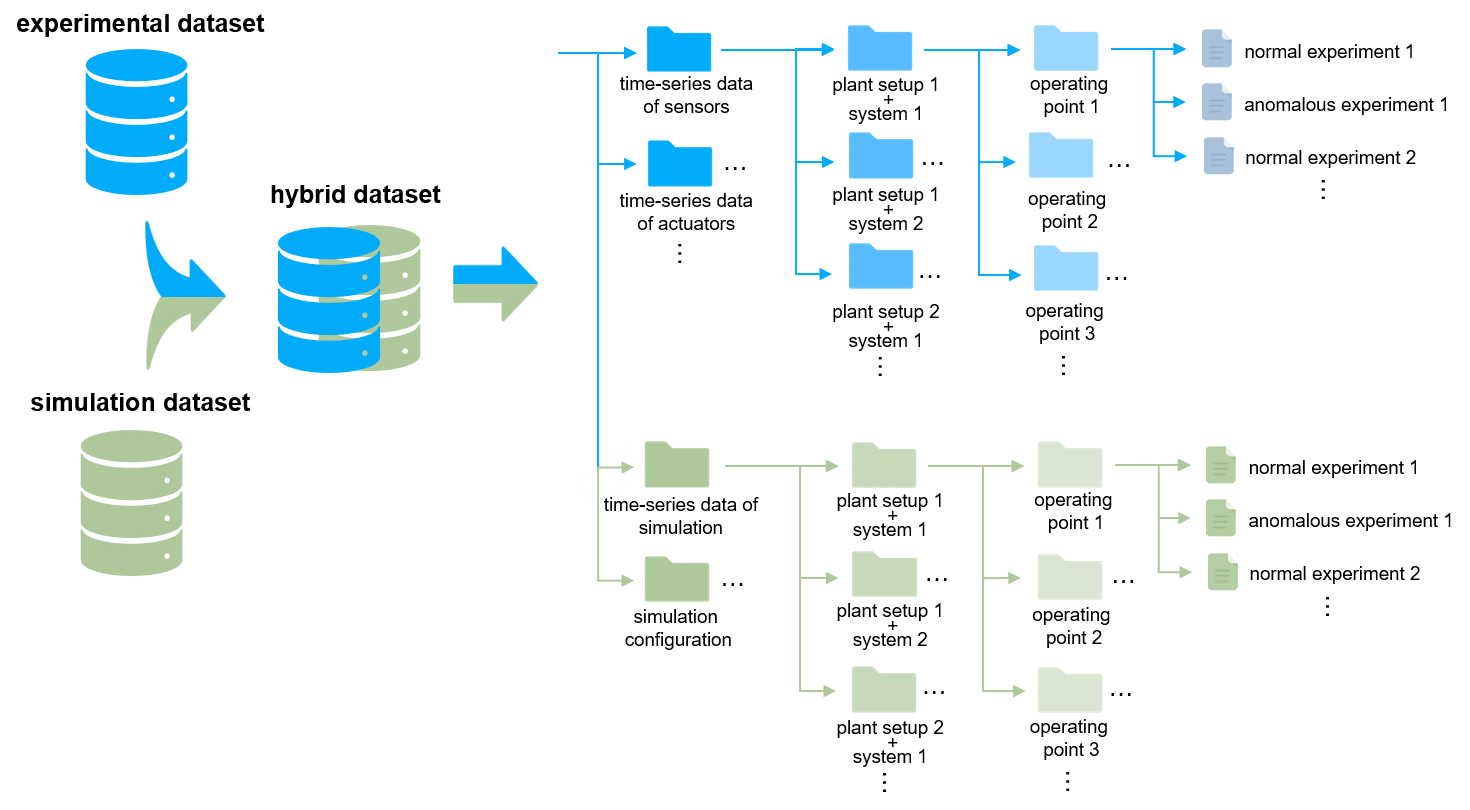}
   \caption{Overview of the extension of the experimental database~\cite{Arweiler2026} with the simulation data from this work. The hierarchy level decreases from left to right. The top-level folders represent the additional modalities of the simulated datasets, which are included in the same main set as the experimental data from~\cite{Arweiler2026}, while the actual data are stored at the lowest level.}
   \label{fig:datastructure}
\end{figure}

The simulation data, including configuration files and comparisons to the experimental data, were added to the batch distillation database introduced in~\cite{Arweiler2026}, creating a large hybrid database of experimental and simulation data. This hybrid database is published on \url{https://zenodo.org/records/19520510}, starting from version 1.1.0, as an extension to the batch distillation database, which is organized as shown in Figure~\ref{fig:datastructure}. 
In the main folder for the batch distillation plant data, all data modalities, including the novel simulation data modalities, are subsumed. On the next lower level, different combinations of plant setups and chemical systems are distinguished; on the second lowest level, operating points are distinguished. The actual simulation configuration files and time-series simulation data can be found on the lowest level of their respective folder tree.

The configuration files consist of two \texttt{.xml} files; one contains all model parameters for pure-component properties, which cannot be made public, and binary interaction \rev{parameters}. The other contains all plant and process control parameters and includes a mapping to the anomaly identifiers from~\cite{Arweiler2026}, indicating whether the respective anomalies of \rev{an} experiment are simulated. \rev{As the anomaly identifiers are unique, they can be used to access detailed descriptions of the respective anomalies.} In the hybrid database\rev{,} we only report the configuration file containing plant and process control parameters, \rev{while the} binary interaction parameters are reported in the Supporting Information.
An example of an \texttt{.xml} configuration file for an anomalous process is provided in the Supporting Information.
\rev{The simulation time-series data are provided in \texttt{.csv} files, with one file per simulated experiment.} \rev{Overall, the hybrid dataset includes the simulation data for 115 experiments, comprising almost 1.2 million time steps in total.}  
\rev{The hybrid dataset contains both fault-free and anomalous experiments. The simulated anomaly types include anomalies introduced by perturbing the control parameter setpoints, covering 36 of the 71 confirmed anomaly events described in~\cite{Arweiler2026}. Moreover, simulations of 37 fault-free experiments are available. For 43 anomalous experiments, the respective anomaly effects could not be represented by perturbing model parameters; for these cases, simulation data for the corresponding unperturbed experiment are provided. All experiments are sorted using the identifiers introduced by~\cite{Arweiler2026}, which enables the simulation data to be matched to the corresponding experimental runs.} \rev{The simulation data only comprise operating data; the start of the simulation is, hence, aligned with the time of the start of the operating phase of the respective distillation experiment. To match the sampling frequency of the experimental data, the simulation data are interpolated to a sampling interval of one second between the temporal collocation points of the simulation model.}
\rev{For each} simulated \rev{experiment,} time-series \rev{data} for liquid molar holdup, vapor up-flow, liquid down-flow, enthalpies and pressures per stage, for liquid mole fractions and vapor mole fractions per stage and component, and for heat duty, efflux ratio, and fluxes of distillate and reflux streams \rev{are reported}. \rev{The complete list of reported variables, including their naming convention and physical units, is provided in the Supporting Information.} The simulation of each experiment extends the distillation process at least as far as the actual experiment; if feasible, the simulation was run until the reboiler vessel was empty. \rev{Further details on the experimental part of the hybrid dataset, including the experimental setup, measured variables, anomaly descriptions, data format and access information, are provided in~\cite{Arweiler2026}.}

\section{Conclusions}
\label{sec:Conclusion}
Large, diverse, and well-annotated datasets are a key prerequisite for training and benchmarking deep anomaly detection (AD) methods for chemical processes. Building on a recently released experimental batch distillation database with structured anomaly annotations, this work adds a consistent simulation layer, thereby creating an openly available hybrid dataset that combines experimental and simulated time-series data.

From a process-simulation perspective, we demonstrate that a laboratory-scale batch distillation campaign can be simulated in a largely automated and physically consistent manner. Using the metadata from the experimental database, initial and operating conditions are automatically retrieved and translated into simulation scenarios. After calibration to a single reference experiment, the simulator predicts the dynamics of the remaining experiments across a wide range of conditions. The developed workflow enables systematic “replay” simulations of confirmed anomalies with known causes. At the same time, open gaps in anomaly simulation remain: we only considered anomalies induced by perturbations that can be modeled straightforwardly, such as set-point changes. The modeling of other anomalies, for example, those caused by foaming, remains an open challenge for future research.

From an AD perspective, the hybrid dataset provides two complementary benefits. First, it extends the experimental database with simulated trajectories for most runs, enabling controlled comparisons between measurements and physically consistent model outputs. Second, it enables the generation of additional pseudo-experimental data at low cost, both by exploring operating conditions and fault scenarios that are impractical or undesirable to realize experimentally and by creating larger and more diverse training corpora than would be feasible from experiments alone. Fully exploiting this potential, however, requires bridging the gap between simulated data and experimental data that contain noise, sensor errors, failures, and various inconsistencies. Machine learning provides techniques for such simulation-to-experiment translation (e.g., style transfer), which can now be systematically tested and refined based on the released hybrid dataset.

Experimental and simulated data overlap only partially: the overlap is valuable for validation and for learning mappings between domains, while the simulation domain additionally provides access to fully specified, internally consistent state and parameter information that is not directly observable in experiments. This combination motivates future work to better leverage simulated data for robust learning-based AD. Overall, the presented simulator and workflow, together with the released hybrid dataset, provide a practical basis for developing, testing, and comparing deep AD methods on chemical-process time series at a scale that is difficult to achieve experimentally.

\section*{Acknowledgements}
The authors gratefully acknowledge funding from the Deutsche Forschungsgemeinschaft (DFG), through the DFG Research Unit FOR 5359 ``Deep Learning on Sparse Chemical Process Data" (grant number 459419731).

\section*{Conflicts of interest}
There are no conflicts of interest to declare.

\section*{Author contributions}
Jennifer Werner: Conceptualization, Software, Validation, Investigation, Data curation, Writing -- original draft, Visualization. 
Justus Arweiler: Conceptualization, Investigation, Validation, Writing -- original draft, Data curation, Visualization.
Indra Jungjohann: Conceptualization, Investigation, Validation, Writing -- original draft, Data curation, Visualization.
Jochen Schmid: Conceptualization, Methodology, Formal analysis, Validation, Investigation,  Writing -- original draft, Supervision.
Fabian Jirasek: Conceptualization,  Writing -- review \& editing, Supervision, Project administration, Funding acquisition. 
Hans Hasse: Conceptualization, Writing -- review \& editing, Supervision, Project administration, Funding acquisition. 
Michael Bortz: Conceptualization, Resources, Writing -- review \& editing, Supervision, Project administration, Funding acquisition. 

\section*{Code availability statement}
As parts of the code were developed in industrial collaborations, it cannot currently be released openly; however, interested users are encouraged to contact the authors to discuss possible forms of access.

\section*{Data availability statement}
The simulation dataset is available in an open Zenodo repository (https://doi.org/10.5281/zenodo.19520510, data modalities \texttt{13\_...\_Timeseries\_Simulation} and \texttt{14\_...\_Simulation\_Configuration}). The data modalities contain the data structured as described in this paper in \texttt{.csv} and \texttt{.xml} files format. When publishing results based on the simulation data (data modalities \texttt{13\_...\_Timeseries\_Simulation}, \texttt{14\_...\_Simulation\_Configuration}), users should cite this paper, and the dataset DOI of the used version.

\newpage

\bibliographystyle{elsarticle-num}
\bibliography{references_V3} 

@Article{WernerSchmid2025,
  author   = {Jennifer Werner and Jochen Schmid and Lorenz T. Biegler and Michael Bortz},
  journal  = {Fluid Phase Equilibr.},
  title    = {An equation-based batch distillation simulation to evaluate the effect of multiplicities in thermodynamic activity coefficients},
  year     = {2025},
  issn     = {0378-3812},
  pages    = {114465},
  volume   = {598},
  doi      = {10.1016/j.fluid.2025.114465},
  fjournal = {Fluid Phase Equilibria},
}

@inproceedings{Gatys2016,
  author={Gatys, Leon A. and Ecker, Alexander S. and Bethge, Matthias},
  booktitle={2016 IEEE Conference on Computer Vision and Pattern Recognition (CVPR)}, 
  title={Image Style Transfer Using Convolutional Neural Networks}, 
  year={2016},
  volume={},
  number={},
  pages={2414-2423},
  doi={10.1109/CVPR.2016.265}}

@inproceedings{Xu2024,
  author={Xu, Xinhe and Wang, Zhuoer and Zhang, Yihan and Liu, Yizhou and Wang, Zhaoyue and Xu, Zhihao and Zhao, Muhan and Luo, Huaiying},
  booktitle={2024 5th International Conference on Big Data \& Artificial Intelligence \& Software Engineering (ICBASE)}, 
  title={Style Transfer: From Stitching to Neural Networks}, 
  year={2024},
  volume={},
  number={},
  pages={526-530},
  doi={10.1109/ICBASE63199.2024.10762296}
}

@inproceedings{El-Laham2022,
author = {El-Laham, Yousef and Vyetrenko, Svitlana},
title = {StyleTime: Style Transfer for Synthetic Time Series Generation},
year = {2022},
isbn = {9781450393768},
publisher = {Association for Computing Machinery},
address = {New York, NY, USA},
doi = {10.1145/3533271.3561772},
booktitle = {Proceedings of the Third ACM International Conference on AI in Finance},
pages = {489–496},
numpages = {8},
location = {New York, NY, USA},
series = {ICAIF '22}
}

@misc{MainDataset,
    title = {Batch Distillation Data for Developing Machine Learning Anomaly Detection Methods},
    author = {Arweiler, J. and Jungjohann, I. and Werner, J. and Muraleedharan, A. and Schmid, J. and Leitte, H. and Burger, J. and Münnemann, K. and Bortz, M. and  Jirasek, F. and Hasse, H.},
    doi ={10.5281/zenodo.17395543},
    url ={https://zenodo.org/records/17395543},
    note={Version 1.1.2},
    date={2026},
    journal={Zenodo}
}

@book{Campbell2019,
    author = {Campbell, Stephen and Ilchmann, Achim and Mehrmann, Volker and Reis, Timo},
    title = {Applications of Differential-Algebraic Equations: Examples and Benchmarks},
    publisher = {Springer},
    year = {2019},
    DOI = {10.1007/978-3-030-03718-5}
}

@article{Arweiler2026,
    author = {Arweiler, Justus and Jungjohann, Indra and Muraleedharan, Aparna and Leitte, Heike and Burger, Jakob and Münnemann, Kerstin and Jirasek, Fabian and Hasse, Hans},
    title = {Batch distillation data for developing machine learning anomaly detection methods},
    journal = {Sci. Data} ,
    year = {2026},
    volume = {13},
    number = {513},
    DOI = {10.1038/s41597-026-07124-3}
}

@article{HeatCapacity_Borosilicate-glass,
    title = {Specific heat capacity of Apiezon N high vacuum grease and of Duran borosilicate glass},
    author = {Schnelle, Walter and Engelhardt, J. and Gmelin, Eberhard},
    journal = {Cryogenics},
    volume = {39},
    number = {3},
    pages = {271-275},
    year = {1999},
    issn = {0011-2275},
    doi = {10.1016/S0011-2275(99)00035-1},
}

@Article{HeatCapacity_14404stainlesssteel,
  author   = {Pichler, Peter and Simonds, Brian and Sowards, Jeffrey and Pottlacher, Gernot},
  journal  = {J. Mater. Sci.},
  title    = {Measurements of thermophysical properties of solid and liquid NIST SRM 316L stainless steel},
  year     = {2020},
  issn     = {1573-4803},
  number   = {9},
  pages    = {4081-4093},
  volume   = {55},
  doi      = {10.1007/s10853-019-04261-6},
  fjournal = {Journal of Materials Science},
}

@Article{wagner2023timesead,
  author  = {Wagner, Dennis and Michels, Tobias and Schulz, Florian CF and Nair, Arjun and Rudolph, Maja and Kloft, Marius},
  journal = {Trans. Mach. Learn. Res.},
  title   = {Timesead: Benchmarking deep multivariate time-series anomaly detection},
  year    = {2023},
  issn    = {2835-8856},
  url     = {https://openreview.net/forum?id=iMmsCI0JsS},
}

@Article{Hartung2023,
  author    = {Hartung, Fabian and Franks, Billy Joe and Michels, Tobias and Wagner, Dennis and Liznerski, Philipp and Reithermann, Steffen and Fellenz, Sophie and Jirasek, Fabian and Rudolph, Maja and Neider, Daniel and Leitte, Heike and Song, Chen and Kloepper, Benjamin and Mandt, Stephan and Bortz, Michael and Burger, Jakob and Hasse, Hans and Kloft, Marius},
  journal   = {Chem. Ing. Tech.},
  title     = {Deep Anomaly Detection on Tennessee Eastman Process Data},
  year      = {2023},
  issn      = {1522-2640},
  month     = apr,
  number    = {7},
  pages     = {1077--1082},
  volume    = {95},
  doi       = {10.1002/cite.202200238},
  fjournal  = {Chemie Ingenieur Technik},
  publisher = {Wiley},
}

@article{noBoom,
    author = {Wagner, Dennis and Hartung, Fabian and Arweiler, Justus and Muraleedharan, Aparna and Jungjohann, Indra and Nair, Arjun and Reithermann, Steffen and Schulz, Ralf and Bortz, Michael and Neider, Daniel and Leitte, Heike and Pfeffinger, Joachim and Mandt, Stephan and Fellenz, Sophie and Katz, Torsten and Jirasek, Fabian and Burger, Jakob and Hasse, Hans and Kloft, Marius},
    title = {NoBOOM: Chemical Process Datasets for Industrial Anomaly Detection},
    journal = {NeurIPS 2025 Datasets and Benchmarks},
    year = {2025} ,
    url={https://openreview.net/forum?id=qiLboR0ocm},
    booktitle={The Thirty-ninth Annual Conference on Neural Information Processing Systems Datasets and Benchmarks Track},
}

@Article{Doherty1978,
  author   = {M.F. Doherty and J.D. Perkins},
  journal  = {Chem. Eng. Sci.},
  title    = {{On the dynamics of distillation processes—I: The simple distillation of multicomponent non-reacting, homogeneous liquid mixtures}},
  year     = {1978},
  issn     = {0009-2509},
  number   = {3},
  pages    = {281-301},
  volume   = {33},
  doi      = {10.1016/0009-2509(78)80086-4},
  fjournal = {Chemical Engineering Science},
}

@Article{VanDongen1985,
  author   = {David B. {Van Dongen} and Michael F. Doherty},
  journal  = {Chem. Eng. Sci.},
  title    = {On the dynamics of distillation processes—VI. batch distillation},
  year     = {1985},
  issn     = {0009-2509},
  number   = {11},
  pages    = {2087-2093},
  volume   = {40},
  doi      = {10.1016/0009-2509(85)87026-3},
  fjournal = {Chemical Engineering Science},
}

@Article{cervantes1998,
  author   = {Cervantes, A. and Biegler, L. T.},
  journal  = {AlChE J.},
  title    = {Large-scale DAE optimization using a simultaneous NLP formulation},
  year     = {1998},
  number   = {5},
  pages    = {1038-1050},
  volume   = {44},
  doi      = {10.1002/aic.690440505},
  eprint   = {https://aiche.onlinelibrary.wiley.com/doi/pdf/10.1002/aic.690440505},
  fjournal = {AIChE Journal},
}

@Article{cervantes2000,
  author   = {Arturo M. Cervantes and Andreas Wächter and Reha H. Tütüncü and Lorenz T. Biegler},
  journal  = {Comput. Chem. Eng.},
  title    = {A reduced space interior point strategy for optimization of differential algebraic systems},
  year     = {2000},
  issn     = {0098-1354},
  number   = {1},
  pages    = {39-51},
  volume   = {24},
  doi      = {10.1016/S0098-1354(00)00302-1},
  fjournal = {Computers & Chemical Engineering},
}

@Article{CERVANTES200041,
  author   = {Arturo M Cervantes and Lorenz T Biegler},
  journal  = {J. Comput. Appl. Math.},
  title    = {A stable elemental decomposition for dynamic process optimization},
  year     = {2000},
  issn     = {0377-0427},
  number   = {1},
  pages    = {41-57},
  volume   = {120},
  doi      = {10.1016/S0377-0427(00)00302-2},
  fjournal = {Journal of Computational and Applied Mathematics},
}

@Article{biegler2002,
  author   = {Lorenz T. Biegler and Arturo M. Cervantes and Andreas Wächter},
  journal  = {Chem. Eng. Sci.},
  title    = {Advances in simultaneous strategies for dynamic process optimization},
  year     = {2002},
  issn     = {0009-2509},
  number   = {4},
  pages    = {575-593},
  volume   = {57},
  doi      = {10.1016/S0009-2509(01)00376-1},
  fjournal = {Chemical Engineering Science},
}

@Article{ragunathan2004,
  author   = {Arvind U Raghunathan and M {Soledad Diaz} and Lorenz T Biegler},
  journal  = {Comput. Chem. Eng.},
  title    = {{An MPEC formulation for dynamic optimization of distillation operations}},
  year     = {2004},
  issn     = {0098-1354},
  note     = {Special Issue for Professor Arthur W. Westerberg},
  number   = {10},
  pages    = {2037-2052},
  volume   = {28},
  doi      = {10.1016/j.compchemeng.2004.03.015},
  fjournal = {Computers & Chemical Engineering},
}

@Article{eckert2008mathematical,
  author    = {Eckert, Egon and Van{\v{e}}k, Tom{\'a}{\v{s}}},
  journal   = {Chem. Pap.},
  title     = {Mathematical modelling of selected characterisation procedures for oil fractions},
  year      = {2008},
  pages     = {26--33},
  volume    = {62},
  fjournal  = {Chemical Papers},
  publisher = {Springer},
}

@Article{lopez2016rigorous,
  author    = {Lopez-Saucedo, Edna Soraya and Grossmann, Ignacio E and Segovia-Hernandez, Juan Gabriel and Hern{\'a}ndez, Salvador},
  journal   = {Chem. Eng. Res. Des.},
  title     = {{Rigorous modeling, simulation and optimization of a conventional and nonconventional batch reactive distillation column: A comparative study of dynamic optimization approaches}},
  year      = {2016},
  pages     = {83--99},
  volume    = {111},
  fjournal  = {Chemical Engineering Research and Design},
  publisher = {Elsevier},
}

@InCollection{bortz2019estimating,
  author    = {Bortz, Michael and Heese, Raoul and Scherrer, Alexander and Gerlach, Thomas and Runowski, Thomas},
  booktitle = {Comput. Aided Chem. Eng.},
  publisher = {Elsevier},
  title     = {Estimating mixture properties from batch distillation using semi-rigorous and rigorous models},
  year      = {2019},
  isbn      = {9780128186343},
  pages     = {643--648},
  volume    = {46},
  doi       = {10.1016/B978-0-12-818634-3.50108-9},
  issn      = {1570-7946},
}

@incollection{Mohring2022,
title = {{Modeling and optimizing dynamic networks: Applications in process engineering and energy supply}},
editor = {Michael Bortz and Norbert Asprion},
booktitle = {Simulation and Optimization in Process Engineering},
publisher = {Elsevier},
pages = {143-160},
year = {2022},
isbn = {978-0-323-85043-8},
doi = {10.1016/B978-0-323-85043-8.00013-1},
author = {Jan Mohring and Jochen Schmid and Jarosław Wlazło and Raoul Heese and Thomas Gerlach and Thomas Kochenburger and Michael Bortz},
}

@Article{qian2023nonlinear,
  author    = {Qian, Xing and Lin, Kuan-Han and Jia, Shengkun and Biegler, Lorenz T and Huang, Kejin},
  journal   = {AlChE J.},
  title     = {Nonlinear model predictive control for dividing wall columns},
  year      = {2023},
  number    = {6},
  pages     = {e18062},
  volume    = {69},
  fjournal  = {AIChE Journal},
  publisher = {Wiley Online Library},
  doi       = {10.1002/aic.18062},
}

@Book{hairer1991ii,
  author    = {Hairer, Ernst and Wanner, Gerhard},
  publisher = {Springer Berlin Heidelberg},
  title     = {Solving ordinary differential equations II: Stiff and differential-algebraic problems},
  year      = {1991},
  isbn      = {9783642052217},
  doi       = {10.1007/978-3-642-05221-7},
  issn      = {0179-3632},
  journal   = {Springer},
}

@Article{Venkatasubramanian2003,
  author    = {Venkatasubramanian, Venkat and Rengaswamy, Raghunathan and Yin, Kewen and Kavuri, Surya N.},
  journal   = {Comput. Chem. Eng.},
  title     = {A review of process fault detection and diagnosis Part I: Quantitative model-based methods},
  year      = {2003},
  issn      = {0098-1354},
  month     = mar,
  number    = {3},
  pages     = {293--311},
  volume    = {27},
  doi       = {10.1016/s0098-1354(02)00160-6},
  fjournal  = {Computers & Chemical Engineering},
  publisher = {Elsevier BV},
}

@Article{Venkatasubramanian2003a,
  author    = {Venkatasubramanian, Venkat and Rengaswamy, Raghunathan and Kavuri, Surya N},
  journal   = {Comput. Chem. Eng.},
  title     = {A review of process fault detection and diagnosis Part II: Qualitative models and search strategies},
  year      = {2003},
  issn      = {0098-1354},
  month     = mar,
  number    = {3},
  pages     = {313--326},
  volume    = {27},
  doi       = {10.1016/s0098-1354(02)00161-8},
  fjournal  = {Computers & Chemical Engineering},
  publisher = {Elsevier BV},
}

@Article{Venkatasubramanian2003b,
  author    = {Venkatasubramanian, Venkat and Rengaswamy, Raghunathan and Kavuri, Surya N. and Yin, Kewen},
  journal   = {Comput. Chem. Eng.},
  title     = {A review of process fault detection and diagnosis Part III: Process history based methods},
  year      = {2003},
  issn      = {0098-1354},
  month     = mar,
  number    = {3},
  pages     = {327--346},
  volume    = {27},
  doi       = {10.1016/s0098-1354(02)00162-x},
  fjournal  = {Computers & Chemical Engineering},
  publisher = {Elsevier BV},
}

@Book{Chiang2001,
  author    = {Chiang, Leo H.},
  publisher = {Springer},
  title     = {Fault Detection and Diagnosis in Industrial Systems},
  year      = {2001},
  address   = {London},
  isbn      = {9781447103479},
  series    = {Advanced Textbooks in Control and Signal Processing},
  pagetotal = {27981},
  ppn_gvk   = {749179139},
}

@Article{Chandola2009,
  author    = {Chandola, Varun and Banerjee, Arindam and Kumar, Vipin},
  journal   = {ACM Comput. Surv.},
  title     = {Anomaly detection: A survey},
  year      = {2009},
  issn      = {1557-7341},
  month     = jul,
  number    = {3},
  pages     = {1--58},
  volume    = {41},
  doi       = {10.1145/1541880.1541882},
  publisher = {Association for Computing Machinery (ACM)},
}

@InProceedings{Chadha2019,
  author    = {Chadha, Gavneet Singh and Rabbani, Arfyan and Schwung, Andreas},
  booktitle = {2019 IEEE 17th International Conference on Industrial Informatics (INDIN)},
  title     = {Comparison of Semi-supervised Deep Neural Networks for Anomaly Detection in Industrial Processes},
  year      = {2019},
  month     = jul,
  pages     = {214--219},
  publisher = {IEEE},
  doi       = {10.1109/indin41052.2019.8972172},
}

@Article{Tian2020,
  author    = {Tian, Wende and Liu, Zijian and Li, Lening and Zhang, Shifa and Li, Chuankun},
  journal   = {Chinese J. Chem. Eng.},
  title     = {Identification of abnormal conditions in high-dimensional chemical process based on feature selection and deep learning},
  year      = {2020},
  issn      = {1004-9541},
  month     = jul,
  number    = {7},
  pages     = {1875--1883},
  volume    = {28},
  doi       = {10.1016/j.cjche.2020.05.003},
  fjournal  = {Chinese Journal of Chemical Engineering},
  publisher = {Elsevier BV},
}

@Article{Wu2024,
  author    = {Wu, Gaochang and Zhang, Yapeng and Deng, Lan and Zhang, Jingxin and Chai, Tianyou},
  title     = {Cross-Modal Learning for Anomaly Detection in Complex Industrial Process: Methodology and Benchmark},
  year      = {2024},
  copyright = {arXiv.org perpetual, non-exclusive license},
  doi       = {10.48550/ARXIV.2406.09016},
  journal  = {arXiv},
}

@InProceedings{Inoue2017,
  author    = {Inoue, Jun and Yamagata, Yoriyuki and Chen, Yuqi and Poskitt, Christopher M. and Sun, Jun},
  booktitle = {2017 IEEE International Conference on Data Mining Workshops (ICDMW)},
  title     = {Anomaly Detection for a Water Treatment System Using Unsupervised Machine Learning},
  year      = {2017},
  month     = nov,
  pages     = {1058--1065},
  publisher = {IEEE},
  doi       = {10.1109/icdmw.2017.149},
}

@InProceedings{Monroy2009,
  author    = {Monroy, Isaac and Escudero, Gerard and Graells, Moisès},
  publisher = {Elsevier},
  title     = {Anomaly detection in batch chemical processes},
  year      = {2009},
  isbn      = {9780444534330},
  booktitle = {19th European Symposium on Computer Aided Process Engineering},
  doi       = {10.1016/s1570-7946(09)70043-4},
  issn      = {1570-7946},
  pages     = {255--260},
}

@Article{Song2019,
  author    = {Song, Bomi and Suh, Yongyoon},
  journal   = {J. Loss Prev. Process Ind.},
  title     = {Narrative texts-based anomaly detection using accident report documents: The case of chemical process safety},
  year      = {2019},
  issn      = {0950-4230},
  month     = jan,
  pages     = {47--54},
  volume    = {57},
  doi       = {10.1016/j.jlp.2018.08.010},
  fjournal  = {Journal of Loss Prevention in the Process Industries},
  publisher = {Elsevier BV},
}

@Book{Russell2000,
  author    = {Russell, Evan L. and Chiang, Leo H. and Braatz, Richard D.},
  publisher = {Springer London},
  title     = {Data-driven Methods for Fault Detection and Diagnosis in Chemical Processes},
  year      = {2000},
  isbn      = {9781447104094},
  doi       = {10.1007/978-1-4471-0409-4},
  fjournal  = {Advances in Industrial Control},
  issn      = {2193-1577},
  journal   = {Adv. Ind. Control},
}

@Article{Schmidl2022,
  author    = {Schmidl, Sebastian and Wenig, Phillip and Papenbrock, Thorsten},
  journal   = {Proceedings of the VLDB Endowment},
  title     = {Anomaly detection in time series: a comprehensive evaluation},
  year      = {2022},
  issn      = {2150-8097},
  month     = may,
  number    = {9},
  pages     = {1779--1797},
  volume    = {15},
  doi       = {10.14778/3538598.3538602},
  publisher = {Association for Computing Machinery (ACM)},
}

@Article{Darban2024,
  author    = {Zamanzadeh Darban, Zahra and Webb, Geoffrey I. and Pan, Shirui and Aggarwal, Charu and Salehi, Mahsa},
  journal   = {ACM Comput. Surv.},
  title     = {Deep Learning for Time Series Anomaly Detection: A Survey},
  year      = {2024},
  issn      = {1557-7341},
  month     = oct,
  number    = {1},
  pages     = {1--42},
  volume    = {57},
  doi       = {10.1145/3691338},
  publisher = {Association for Computing Machinery (ACM)},
}

@article{Muraleedharan2025,
  author  = {Muraleedharan, Aparna and Ferre, Alvaro and Arweiler, Justus and Jungjohann, Indra and Jirasek, Fabian and Hasse, Hans and Burger, Jakob},
  title   = {Experimental time series data with and without anomalies from a continuous distillation mini-plant for development of machine learning anomaly detection methods},
  year    = {2025},
  doi     = {10.31224/5631},
  url     = {https://engrxiv.org/preprint/view/5631},
  note    = {engrXiv preprint}
}

@Article{Downs1993,
  author    = {Downs, J.J. and Vogel, E.F.},
  journal   = {Comput. Chem. Eng.},
  title     = {A plant-wide industrial process control problem},
  year      = {1993},
  issn      = {0098-1354},
  month     = mar,
  number    = {3},
  pages     = {245--255},
  volume    = {17},
  doi       = {10.1016/0098-1354(93)80018-i},
  fjournal  = {Computers & Chemical Engineering},
  publisher = {Elsevier BV},
}

@Article{Rieth2017,
  author    = {Rieth, Cory A. and Amsel, Ben D. and Tran, Randy and Cook, Maia B.},
  title     = {Additional Tennessee Eastman Process Simulation Data for Anomaly Detection Evaluation},
  year      = {2017},
  doi       = {10.7910/DVN/6C3JR1},
  journal = {Harvard Dataverse},
}

@Article{Park2020,
AUTHOR = {Park, You-Jin and Fan, Shu-Kai S. and Hsu, Chia-Yu},
TITLE = {A Review on Fault Detection and Process Diagnostics in Industrial Processes},
JOURNAL = {Processes},
VOLUME = {8},
YEAR = {2020},
NUMBER = {9},
ARTICLE-NUMBER = {1123},
ISSN = {2227-9717},
DOI = {10.3390/pr8091123}
}

@Article{Ji2022,
AUTHOR = {Ji, Cheng and Sun, Wei},
TITLE = {A Review on Data-Driven Process Monitoring Methods: Characterization and Mining of Industrial Data},
JOURNAL = {Processes},
VOLUME = {10},
YEAR = {2022},
NUMBER = {2},
ARTICLE-NUMBER = {335},
ISSN = {2227-9717},
DOI = {10.3390/pr10020335}
}

@Article{Lopez2016,
  author   = {Edna Soraya Lopez-Saucedo and Ignacio E. Grossmann and Juan Gabriel Segovia-Hernandez and Salvador Hernández},
  journal  = {Chem. Eng. Res. Des.},
  title    = {Rigorous modeling, simulation and optimization of a conventional and nonconventional batch reactive distillation column: A comparative study of dynamic optimization approaches},
  year     = {2016},
  issn     = {0263-8762},
  pages    = {83-99},
  volume   = {111},
  doi      = {10.1016/j.cherd.2016.04.005},
  fjournal = {Chemical Engineering Research and Design},
}

@conference{Gruetzmann2006,
    author = {Gruetzmann, Sven and Kapala, Thomas and Fieg, Georg},
    booktitle = {16th European Symposium of Computer Aided Process Engineering and 9th International Symposium on Process Systems Engineering},
    title = {Dynamic Modelling of Complex Batch Distillation
Starting from Ambient Conditions},
    year = {2006}
}

@book{ascher1998computer,
  title={Computer methods for ordinary differential equations and differential-algebraic equations},
  author={Ascher, Uri M and Petzold, Linda R},
  year={1998},
  publisher={SIAM}
}

@book{kunkel2006differential,
  title={Differential-algebraic equations: analysis and numerical solution},
  author={Kunkel, Peter},
  volume={2},
  year={2006},
  publisher={European Mathematical Society}
}

@Article{bachmann1990methods,
  author    = {Bachmann, R and Br{\"u}ll, L and Mrziglod, Th and Pallaske, U},
  journal   = {Comput. Chem. Eng.},
  title     = {On methods for reducing the index of differential algebraic equations},
  year      = {1990},
  number    = {11},
  pages     = {1271--1273},
  volume    = {14},
  doi       = {10.1016/0098-1354(90)80007-X},
  fjournal  = {Computers & Chemical Engineering},
  publisher = {Elsevier},
}

@misc{Aspen,
    title     = {{Aspen Plus\textregistered, Version 15}},
    publisher = {{Aspen Technology, Inc.}},
    year      = {2025},
    address   = {Bedford, MA, USA},
    note      = {Process simulation software}
}

@Book{AspenTextbook,
  author    = {Haydary, Juma},
  publisher = {Wiley},
  title     = {Chemical Process Design and Simulation: Aspen Plus and Aspen Hysys Applications},
  year      = {2018},
  isbn      = {9781119311478},
  month     = nov,
  doi       = {10.1002/9781119311478},
}

@article{nagda2025diffstylets,
  title         = {DiffStyleTS: Diffusion Model for Style Transfer in Time Series},
  author        = {Nagda, Mayank and Ostheimer, Phil and Arweiler, Justus and Jungjohann, Indra and Werner, Jennifer and Wagner, Dennis and Muraleedharan, Aparna and Jafari, Pouya and Schmid, Jochen and Jirasek, Fabian and Burger, Jakob and Bortz, Michael and Hasse, Hans and Mandt, Stephan and Kloft, Marius and Fellenz, Sophie},
  year          = {2025},
  eprint        = {2510.11335},
  archivePrefix = {arXiv},
  primaryClass  = {cs.LG},
  doi           = {10.48550/arXiv.2510.11335},
}

\end{document}